\documentclass[lettersize,journal]{IEEEtran}
\usepackage{amsmath,amsfonts}
\usepackage{algorithmic}
\usepackage{algorithm}
\usepackage{array}
\usepackage[caption=false,font=normalsize,labelfont=sf,textfont=sf]{subfig}
\usepackage{textcomp}
\usepackage{stfloats}
\usepackage{url}
\usepackage{verbatim}
\usepackage{graphicx}
\usepackage{cite}
\usepackage[colorlinks=true, linkcolor=blue, citecolor=blue, urlcolor=blue]{hyperref}
\usepackage{booktabs}
\usepackage{multirow}

\usepackage{amssymb}

\begin{document}

\title{SelaVPR++: Towards Seamless Adaptation of Foundation Models for Efficient Place Recognition}

\author{Feng Lu, Tong Jin, Xiangyuan Lan, Lijun Zhang, Yunpeng Liu, Yaowei Wang, Chun Yuan,~\IEEEmembership{Senior Member,~IEEE}
\thanks{Feng Lu and Chun Yuan are with the Tsinghua Shenzhen International Graduate School, Tsinghua University, Shenzhen, China, and also with the Pengcheng Laboratory, Shenzhen, China (e-mail: \{lf22@mails, yuanc@sz\}.tsinghua.edu.cn).}
\thanks{Tong Jin and Yunpeng Liu are with the Shenyang Institute of Automation, Chinese Academy of Sciences, Shenyang, China
.}
\thanks{Xiangyuan Lan and Yaowei Wang are with the Pengcheng Laboratory, Shenzhen, China, and Xiangyuan Lan is also with the Pazhou Laboratory (Huangpu), Guangzhou, China
.}
\thanks{Lijun Zhang is with the Chongqing Institute of Green and Intelligent Technology, Chinese Academy of Sciences, Chongqing, China
.}
\thanks{Feng Lu and Tong Jin contributed equally to this paper.}
\thanks{(Corresponding authors: Xiangyuan Lan; Chun Yuan.)}
}

\markboth{Journal of \LaTeX\ Class Files,~Vol.~14, No.~8, August~2021}%
{Shell \MakeLowercase{\textit{et al.}}: A Sample Article Using IEEEtran.cls for IEEE Journals}


\maketitle

\begin{abstract}
Recent studies show that the visual place recognition (VPR) method using pre-trained visual foundation models can achieve promising performance.
In our previous work, we propose a novel method to realize seamless adaptation of foundation models to VPR (SelaVPR). This method can produce both global and local features that focus on discriminative landmarks to recognize places for two-stage VPR by a parameter-efficient adaptation approach. Although SelaVPR has achieved competitive results, we argue that the previous adaptation is inefficient in training time and GPU memory usage, and the re-ranking paradigm is also costly in retrieval latency and storage usage. In pursuit of higher efficiency and better performance, we propose an extension of the SelaVPR, called SelaVPR++. 
Concretely, we first design a parameter-, time-, and memory-efficient adaptation method that uses lightweight multi-scale convolution (MultiConv) adapters to refine intermediate features from the frozen foundation backbone. This adaptation method does not back-propagate gradients through the backbone during training, and the MultiConv adapter facilitates feature interactions along the spatial axes and introduces proper local priors, thus achieving higher efficiency and better performance. Moreover, we propose an innovative re-ranking paradigm for more efficient VPR. Instead of relying on local features for re-ranking, which incurs huge overhead in latency and storage, we employ compact binary features for initial retrieval and robust floating-point (global) features for re-ranking. To obtain such binary features, we propose a similarity-constrained deep hashing method, which can be easily integrated into the VPR pipeline. Finally, we improve our training strategy and unify the training protocol of several common training datasets to merge them for better training of VPR models. Extensive experiments show that SelaVPR++ is highly efficient in training time, GPU memory usage, and retrieval latency (6000× faster than TransVPR), as well as outperforms the state-of-the-art methods by a large margin (ranks 1st on MSLS challenge leaderboard). Code and models will be released (and merged with SelaVPR) at \small{\url{https://github.com/Lu-Feng/SelaVPR}}.
\end{abstract}


\section{Introduction}
\IEEEPARstart{V}{isual} place recognition (VPR), also known as image localization \cite{liu2019} or visual geo-localization \cite{benchmark}, aims at coarsely estimating the location of a query place image by searching for its best match from a database of geo-tagged images. VPR has long been studied in robotics and computer vision communities, motivated by its wide applications in mobile robot localization \cite{robotloc} and augmented reality \cite{arloc}, etc. The main challenges of the VPR task include condition (e.g., illumination and weather) changes, viewpoint changes, and perceptual aliasing \cite{survey} (hard to differentiate similar images from different places).

The VPR task is typically addressed by using image retrieval and matching approaches \cite{netvlad,delg} with global or/and local descriptors to represent images. The aggregation algorithms like VLAD \cite{vlad2010,VLAD1,VLAD2} are usually used to aggregate/pool local features into a vector as the global feature. Such global descriptors facilitate fast place retrieval and are robust against viewpoint variations. However, these global features neglect spatial information, making VPR methods based on them prone to perceptual aliasing. A promising solution \cite{delg,patchvlad,transvpr,selavpr}, i.e., two-stage VPR, is to retrieve top-k candidate results in the database using global features, then re-rank these candidates by matching local features. Moreover, VPR model training follows the ``pre-training then fine-tuning" paradigm. Most VPR models are initialized using model parameters pre-trained on ImageNet \cite{ImageNet} and fine-tuned on the VPR datasets, such as Pitts30k \cite{pitts} and GSV-Cities \cite{gsv}. As models and training datasets continue to expand, the training becomes increasingly costly in both computation and memory footprint.

\begin{figure*}[!t]
    \centering
    \includegraphics[width=0.95\linewidth]{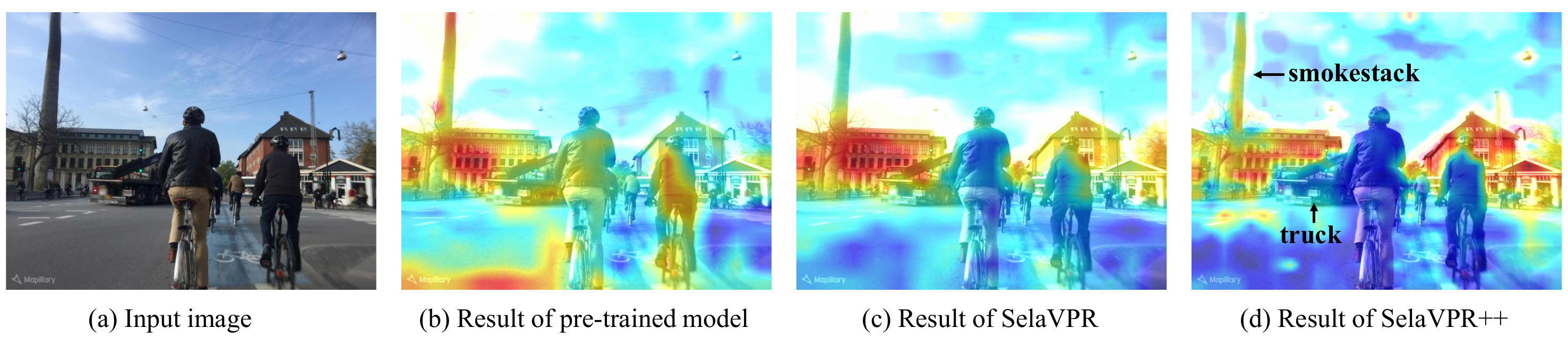}
    \vspace{-0.3cm}
    \caption{
		Heatmap visualizations of feature maps from the pre-trained foundation model, SelaVPR, and SelaVPR++. The pre-trained model pays attention to some regions that are useless for VPR, e.g., dynamic riders. SelaVPR and SelaVPR++ focus on discriminative regions (buildings and trees). Compared with SelaVPR, SelaVPR++ focuses on more landmarks (e.g., smokestack) and eliminates more dynamic interference (e.g., truck), i.e., performs better in detail.}
    \vspace{-0.25cm}
    \label{intro_fig}
\end{figure*}

Recently, foundation models \cite{clip,florence,dinov2} have achieved remarkable performance on many computer vision tasks given their ability to produce well-generalized representations. However, the place image representation produced by the pre-trained model is susceptible to useless (even harmful) dynamic objects (e.g., pedestrians and vehicles), and tends to ignore some static discriminative backgrounds (e.g., buildings and vegetation), as shown in Fig. \ref{intro_fig}. A robust VPR model should focus on the static discriminative landmarks \cite{landmarks2} rather than the dynamic foreground. This results in a gap between the tasks of model pre-training and VPR. Meanwhile, full fine-tuning the foundation model on downstream datasets might forget previously learned knowledge and damage the excellent transferability, i.e., catastrophic forgetting. Additionally, it also results in a large number of trainable parameters and high computation cost. An effective method to address these issues is parameter-efficient transfer learning (PETL) \cite{adapter,prompttuning}, which has not received wide attention in VPR. 

In our previous ICLR 2024 conference version work \cite{selavpr}, we propose a novel method to realize \textbf{Se}am\textbf{l}ess \textbf{a}daptation of pre-trained foundation models for the VPR task, named \textbf{SelaVPR}. By adding tunable lightweight adapters to the frozen pre-trained model, we achieve a hybrid global-local adaptation method to get both global features and dense local features for two-stage VPR, with the former for retrieving candidate images and the latter for re-ranking. Specifically, the global adaptation is achieved by adding adapters after the MHA layer and in parallel to the MLP layer in each transformer block. The local adaptation is implemented by adding up-convolutional layers after the entire transformer backbone to upsample the feature map. Additionally, we propose a mutual nearest neighbor local feature loss, which can be combined with the commonly used triplet loss to optimize the network. The SelaVPR feature representation can focus on the discriminative landmarks, which is critical to identifying places. Furthermore, we can directly match the local features without spatial verification. As a result, SelaVPR outperforms previous methods on several datasets and only consumes less than 3\%  retrieval time of the mainstream two-stage methods. 

\begin{figure}[!t]
    \centering
    \includegraphics[width=1.\linewidth]{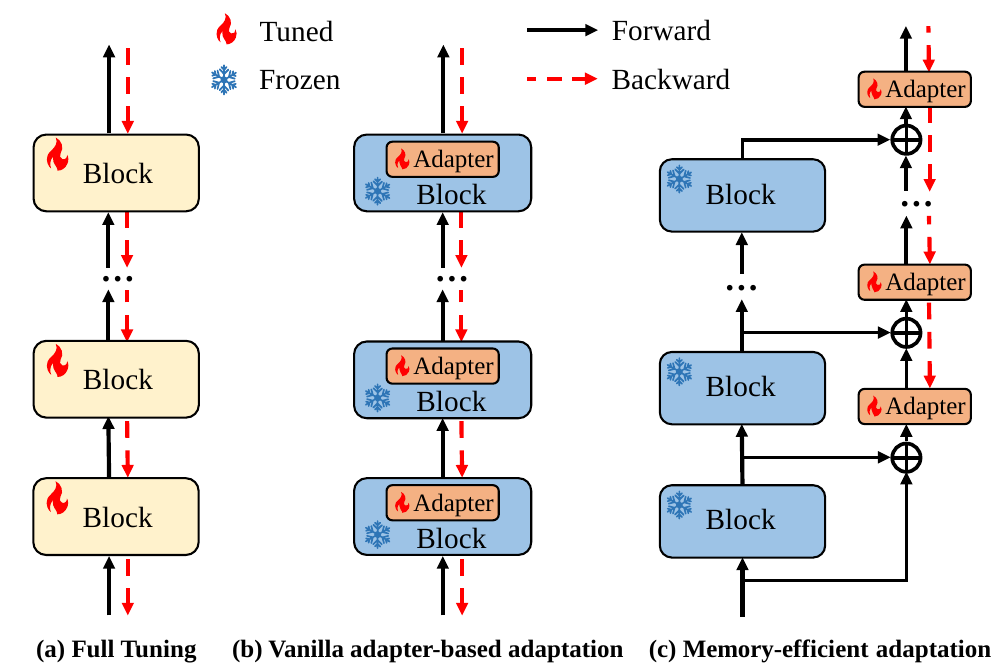}
    \vspace{-0.5cm}
    \caption{
        Comparison between different transfer learning methods. (a) is the common full fine-tuning, in which all blocks are trainable. (b) is a popular PETL method, where only inner adapters are trainable. But the backpropagation still passes through the entire frozen backbone. (c) is the memory-efficient adaptation following the basic framework of previous work \cite{lst}, which reduces training memory usage by eliminating the need for backpropagation through the backbone.
		}
    \vspace{-0.4cm}
    \label{Grad_back}
\end{figure}

Nevertheless, SelaVPR can still be further improved in terms of efficiency and performance. First, although adding the tunable lightweight adapters to the frozen pre-trained model achieves parameter-efficient adaptation, it is not efficient in training time and GPU memory usage. This is because the gradient computation for trainable parameters still requires backpropagation through the frozen backbone, as shown in Fig. \ref{Grad_back} (b), which results in substantial demands on training time and GPU memory. Second, the re-ranking based on dense local features typically incurs huge overhead in time, memory, and storage. Although SelaVPR significantly reduces the retrieval latency compared to common two-stage methods by eliminating the geometric verification requirement, it still requires huge storage to save local features. This limits its application in resource-constrained and large-scale VPR scenes. Additionally, with the development of one-stage VPR methods \cite{cricavpr,salad,salad-cm,boq}, especially training VPR models on large-scale datasets with full supervision \cite{gsv,boq} (instead of weakly supervised training as in SelaVPR), using only global features can also achieve relatively promising results, implying that the paradigm of using local features for re-ranking, which consumes significant resources but offers limited performance gains, is not economical.

To address these issues, we further improve our SelaVPR, and demonstrate a novel insight towards seamless adaptation of foundation models for the more effective and efficient VPR model designs, named \textbf{SelaVPR++}. To begin with, instead of inserting the adapters within the backbone, we connect them to the backbone in parallel and use them to progressively refine the intermediate features from the transformer blocks of the frozen backbone following the basic framework in the work \cite{lst}, as shown in Fig. \ref{Grad_back} (c). This can avoid the gradient backpropagation through the frozen backbone, thus significantly reducing computation and lowering GPU memory footprint during training. In addition, to incorporate local priors beneficial for VPR and make up for the shortcoming of the vanilla adapter, which operates solely on the channel dimension of intermediate features without modeling interactions along the spatial axes, we introduce the multi-scale convolution (MultiConv) adapter to upgrade the adapter used in SelaVPR. Then, we propose an innovative two-stage paradigm for more efficient VPR. Rather than relying on local features for time- and memory-consuming re-ranking as existing methods do, we use compact low-dimensional (e.g., 512-dim) binary features for initial retrieval to get top-k candidate images and robust high-dimensional (e.g., 4096-dim) floating-point features for re-ranking. That is, our method employs global features for both stages, without any local features. Specifically, we utilize deep hashing to obtain compact binary features, in which we design a similarity-constrained quantization loss that only preserves the similarity consistency of feature pairs before and after quantization to address the conflict between common quantization loss and metric loss. We also combine it with straight-through estimation \cite{ste}, which can tackle the gradient issue of quantization operation, for better performance. Since our method computes the Hamming distance between binary features to provide candidates, it can achieve faster retrieval speed than current one-stage methods. Finally, we improve the training strategy and unify the training protocol of several commonly used VPR training datasets. Unlike training on individual datasets with weak supervision in SelaVPR, we combine several datasets in the unified protocol for fully supervised training. Our SelaVPR++ work brings the following main \textbf{contributions} (to the previous SelaVPR):

\textbf{1)} Inspired by the previous memory-efficient adaptation paradigm \cite{lst,metl}, we propose a novel method to achieve seamless and efficient adaptation of foundation models for VPR, in which trainable lightweight MultiConv adapters are used to refine intermediate features from the frozen foundation model (without the gradient backpropagation through it in training), thereby realizing parameter-, time- and memory-efficient adaptation. Meanwhile, the MultiConv adapter not only models spatial interactions between patch tokens but also introduces multi-scale local priors to the VPR model.

\textbf{2)} We design an innovative two-stage VPR paradigm for more efficient VPR, in which we employ compact low-dimensional binary features for initial retrieval and robust high-dimensional floating-point features for re-ranking. It can greatly enhance the retrieval efficiency, e.g., more than 6000× faster than TransVPR \cite{transvpr} and 60× faster than direct retrieval with 4096-dim global descriptors (on the Pitts30k dataset).

\textbf{3)} For achieving deep hashing to get binary place features, we introduce a similarity-constrained quantization loss to avoid the conflict with metric loss, and also combine it with the straight-through estimation method to get better performance.

\textbf{4)} We improve the training strategy and unify the training protocol of commonly used VPR datasets (GSV-Cities \cite{gsv}, SF-XL \cite{cosplace}, Pitts30k \cite{pitts}, and MSLS \cite{msls}) by place category division. Thus, we can merge them for better model training.

\textbf{5)} Extensive results show that our SelaVPR++ significantly outperforms SelaVPR using less training time and memory usage, and also surpasses other SOTA methods with superior efficiency (and ranks 1st on MSLS challenge leaderboard).

\section{Related Work}
\subsection{One-Stage VPR}
The most common VPR approach performs nearest neighbor search using global features to find the most similar place images without considering re-ranking, which is also recognized as one-stage (i.e., global retrieval) VPR. In the early VPR methods, the global features were commonly produced using aggregation algorithms, such as Bag of Words \cite{BoW} and VLAD \cite{vlad2010}, to process the traditional hand-crafted features (e.g., SIFT \cite{SIFT}, SURF \cite{SURF}). With the advancement of deep learning techniques, many works \cite{sunderhaufIROS2015, crn, SPED, landmarks2, semantic,categorization1, categorization2, landmarks3, yin2019, sta-vpr, gcl, gcl2, mixvpr,cricavpr} have employed various deep features for the VPR task. For instance, some works integrated the aggregation methods into neural networks \cite{netvlad,speNetvlad,attentionVLAD} and improved training strategies \cite{sfrs,cosplace,gsv} to achieve better performance, and other studies \cite{binaryNN,quantizedNN} explored model quantization to reduce memory consumption and latency during deep feature extraction. Nevertheless, most one-stage VPR approaches are susceptible to perceptual aliasing due to the use of aggregated features while neglecting spatial information. One recent work \cite{anyloc} first used pre-trained foundation models for the VPR task. However, this work did not perform any fine-tuning, making it difficult to fully unleash the capability of these models for VPR. Some follow-up works \cite{salad,boq} attempted to directly fine-tune (the last multiple blocks of) foundation models, but this way is primarily suitable for medium-size models, where performance may decrease as the model size increases, i.e. hard to train large models \cite{salad}.

\subsection{Two-Stage VPR}
\label{sec:2stagevpr}
The two-stage (i.e., hierarchical) VPR methods with re-ranking \cite{hvpr,seqnet,patchvlad,geowarp,tcl,transvpr,aanet,structvpr,r2former,selavpr,dhevpr} have been proven as an effective way to further improve performance. These approaches typically retrieved top-k candidate images over the whole database using compact global feature representation, such as NetVLAD \cite{netvlad} or Generalized Mean (GeM) pooling \cite{gem}, then re-ranked candidates by performing local matching between the query image and each candidate using local descriptors. However, most of these methods required geometric consistency verification after local matching \cite{delg,patchvlad,transvpr,dhevpr} or taking into account spatial constraints during matching \cite{geowarp,aanet,r2former}, which greatly increases the computational latency. In our SelaVPR \cite{selavpr} work, we fine-tuned a foundation model with a local adaptation module to obtain dense local features, which can be directly used in cross-matching for re-ranking, without time-consuming geometric verification. However, SelaVPR still required substantial storage (and memory) for local features in re-ranking as previous methods, restricting its applicability in resource-constrained and large-scale VPR scenarios. Additionally, with the rapid development of one-stage VPR \cite{cricavpr,salad,boq}, spending substantial time and resources on re-ranking with local features for very limited performance gains has become inefficient and uneconomical. So, in this work, we attempt to develop an innovative two-stage VPR paradigm that performs initial retrieval using compact low-dimensional binary features derived from deep hashing, and then utilizes high-dimensional floating-point features for re-ranking, significantly reducing time, memory, and storage consumption. Although a few previous works \cite{garg2020fast,hashingforVPR1,hashingforVPR2,hashingremote} have used hashing or deep hashing in VPR and similar tasks, they were either not end-to-end trained, or used the sigmoid function to approximate the binarization during training, which is hard to train and has been replaced by the method that imposes the quantization loss (a regularizer) \cite{hashing2016}. More importantly, they used binary features for direct retrieval instead of candidate retrieval with re-ranking, which did not achieve good performance. To our best knowledge, we are the first to apply binary descriptors for initial retrieval in two-stage VPR.

\subsection{Parameter-Efficient Transfer Learning}
Recent work \cite{clip,dino,wang2023image,dinov2} demonstrated that visual foundation models can produce powerful feature representations and achieve excellent performance on multiple tasks. These works commonly trained the ViT \cite{vit} model or its variants with large quantities of parameters on huge amounts of data. The parameter-efficient transfer learning (PETL) \cite{adapter}, a.k.a. parameter-efficient fine-tuning (PEFT), first proposed in natural language processing, is an effective way to adapt foundation models to various downstream tasks, which can reduce trainable parameters and avoid catastrophic forgetting. The main PETL methods fall broadly into three categories: adding task-specific adapters \cite{adapter}, prompt tuning \cite{prompttuning}, and Low-Rank Adaptation (LoRA) \cite{lora}. Our previous works \cite{selavpr,cricavpr} followed the first to adapt the pre-trained foundation models to the VPR task, while they are efficient only in terms of parameters, not in training time and GPU memory usage. Fortunately, there are some recent works \cite{lst,metl} that have preliminarily attempted to explore memory-efficient transfer learning methods. Inspired by them, we design a memory-efficient adaptation architecture tailored for the VPR task. Concretely, we use the tunable adapters to refine the intermediate features from the frozen backbone instead of inserting them into it to avoid backpropagation through the backbone, and upgrade vanilla adapters to MultiConv adapters to introduce the local priors and facilitate feature interactions along the spatial axes. As a result, we can achieve parameter-, time-, and memory-efficient adaptation and get highly robust VPR models.
\begin{figure*}[!t]
    \centering
    \includegraphics[width=0.92\linewidth]{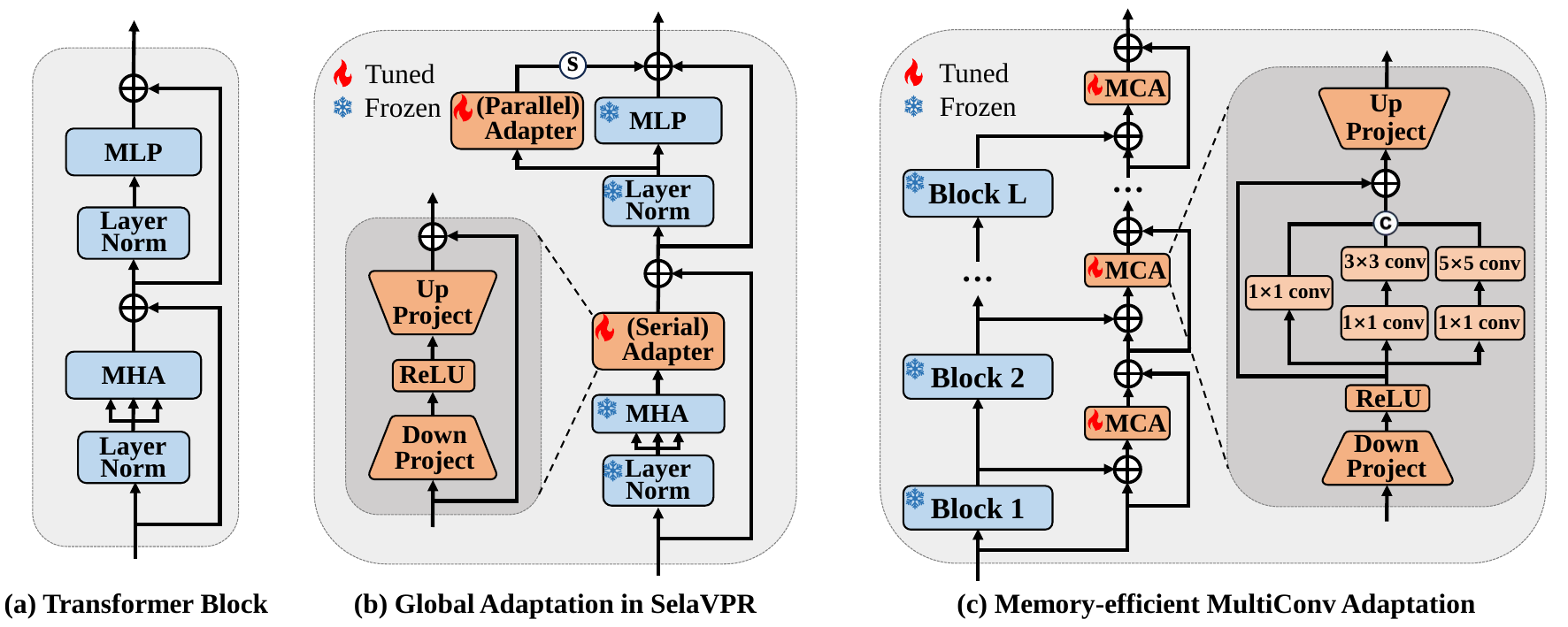}
    \vspace{-0.3cm}
    \caption{
        Illustration of the difference between our memory-efficient MultiConv adaptation network, i.e. (c), and the global adaptation in SelaVPR, i.e. (b). (a) is a transformer block in ViT. Instead of inserting the adapter into the block as (b), we train a parallel side adaptation network as (c), which consists of a series of MultiConv adapters (abbreviated as MCA) to progressively refine the intermediate features from the transformer blocks of the frozen backbone.
		}
    \vspace{-0.3cm}
    \label{MultiConv}
\end{figure*}
\section{Method}
This section describes the proposed SelaVPR++ for two-stage VPR. We first introduce ViT and its use to produce place image representation. Then, we propose the memory-efficient MultiConv adaptation, efficient two-stage VPR paradigm, and similarity-constrained deep hashing. Finally, we present the training strategy and unified dataset for training our model.

\subsection{Preliminary}
\label{Preliminary}
The Vision Transformer (ViT) \cite{vit} and its variants have proven to be powerful for a variety of computer vision tasks including VPR. In this work, we adapt the ViT-based pre-trained foundation model for VPR, so here we give a brief overview of ViT. 

Given an input image, ViT first slices it into $N$ patches and linearly projects them to $D$-dim patch embeddings $x_p\in \mathcal{R}^{N \times D}$, then prepends a learnable \verb|[class]| token to $x_p$ as $x_0=[x_{class};x_p]\in \mathcal{R}^{(N+1) \times D}$. After adding positional embeddings to preserve the positional information, $x_0$ is fed into a series of transformer blocks (i.e., encoder layers) to produce the feature representation.
As shown in Fig. \ref{MultiConv} (a), a transformer block in ViT mainly includes Multi-Head Attention (MHA) and Multi-Layer Perceptron (MLP) modules, as well as Layer Normalization (LN). For the input token sequence, its change process passing through a transformer block is: The MHA is first applied to compute attentional features, then MLP is utilized to realize the feature nonlinearization and dimension transformation. It is formulated as:
\begin{equation}
    x^{\prime}_{l}=\text{MHA}(\text{LN}(x_{l-1}))+x_{l-1},
\end{equation}
\begin{equation}
		x_{l}=\text{MLP}(\text{LN}(x^{\prime}_l))+x^{\prime}_l,
\end{equation}
where $x_{l-1}$ and $x_{l}$ are the output of the $(l-1)$-th and $l$-th transformer block.

For the feature map output by CNN models, a common practice of the VPR method is to use NetVLAD \cite{netvlad} or GeM pooling \cite{gem} to aggregate it into a global feature to conduct nearest neighbor search (i.e., global retrieval). For the ViT model, the output consists of one class token and $N$ patch tokens, where the class token can be directly used as the global feature to represent places \cite{benchmark,r2former}. Meanwhile, $N$ patch tokens can also be reshaped as a feature map (similar to CNN). In this work, instead of using class token, we transform the feature map into the global feature using GeM pooling.

\subsection{Memory-Efficient MultiConv Adaptation}
\label{sec:adaptation}
Although pre-trained foundation models are capable of powerful feature representation, direct use of them in VPR cannot fully unleash their capability due to the gap between the pre-training and VPR tasks. To address it, in our SelaVPR work \cite{selavpr}, we introduced a global adaptation to adapt the pre-trained model so that the feature representation can focus on the static discriminative regions that are beneficial to VPR.
Inspired by previous adapter-based parameter-efficient fine-tuning works \cite{adapter,adaptformer,aim}, our SelaVPR designs the global adaptation as shown in Fig. \ref{MultiConv} (b). Specifically, it adds two adapters in each transformer block. Each adapter is a bottleneck module, which first uses the fully-connected layer to down-project the input to a smaller dimension, then applies a ReLU activation and up-projects it back to the original dimension. The first adapter is a serial adapter that is added after the MHA layer and has a skip-connection internally. The second adapter is a parallel adapter (without skip-connection) connected in parallel to the MLP layer multiplied by a scaling factor $s$. The computation of each global adapted transformer block can be denoted as:
\begin{equation}
            x^{\prime}_{l}=\text{Adapter1}(\text{MHA}(\text{LN}(x_{l-1})))+x_{l-1},
\end{equation}
\begin{equation}
		x_{l}=\text{MLP}(\text{LN}(x^{\prime}_l))+s\cdot \text{Adapter2}(\text{LN}(x^{\prime}_l))+x^{\prime}_l.
\label{eq:adapter}
\end{equation}
However, due to the adapters inserted in the transformer block, this will lead to significant GPU memory consumption during training. Here, we conduct a brief equation derivation and theoretical analysis. Consider a transformer network with $L$ blocks, each containing the frozen parameters $\theta_i$ (frozen modules) and trainable parameters $\alpha_i$ (trainable modules), and producing the output $x_i$, which depends on both parameters $\theta_i, \alpha_i$. What we need to do is to minimize a loss function $\mathcal{L}$ based on stochastic gradient descent. To be specific, the gradient for the trainable parameters $\alpha_i$ is calculated using the chain rule during backpropagation, formulated as follows: 
\begin{equation}
\frac{\partial \mathcal{L}}{\partial \alpha_i} = \frac{\partial \mathcal{L}}{\partial x_i} \frac{\partial x_i}{\partial \alpha_i} = \frac{\partial \mathcal{L}}{\partial x_L} \frac{\partial x_L}{\partial x_{L-1}} \cdots \frac{\partial x_{i+1}}{\partial x_i} \frac{\partial x_i}{\partial \alpha_i}.
\label{chain rule}
\end{equation}
Observing Eq. (\ref{chain rule}), we can find that the gradient of the trainable parameters $\alpha_i$ depends on the gradient with respect to the outputs from subsequent blocks. Although SelaVPR freezes other modules and only tunes the built-in adapters, in order to compute $\frac{\partial \mathcal{L}}{\partial \alpha_i}$, it is inevitable to calculate the gradients of subsequent blocks \{$\frac{\partial \mathcal{L}}{\partial x_L}, \frac{\partial x_L}{\partial x_{L-1}}, \cdots ,\frac{\partial x_{i+1}}{\partial x_i}$\}. This will result in a huge memory overhead. 
 
To this end, inspired by previous works \cite{lst,metl}, we introduce a distinctive adaptation method. Instead of inserting the adapters into the backbone, we use a parallel network that does not require gradient backpropagation through the backbone during training. It is a simple yet effective architecture first proposed in previous work \cite{lst}, which directly takes the intermediate features from the frozen backbone as input and progressively refines them (using lightweight transformer in \cite{lst} but MultiConv adapters in ours) for better representation. Concretely, we adapt the ViT-based foundation model DINOv2 \cite{dinov2} as the backbone, which is composed of a patch embedding layer and $L$ transformer blocks, and thereby $L+1$ intermediate features, $x_0,x_1,x_2...,x_L$, each consisting of $N+1$ tokens with a dimension of $D$, so $x_i \in \mathcal{R}^{(N+1) \times D}$ as mentioned in \ref{Preliminary}. To refine them progressively, we construct the parallel adapters alongside the backbone. Meanwhile, we add residual connections to each adapter for better performance. We represent the output of the $l$-th parallel adapter as $y_l$. The whole computation can be formulated as:
\begin{equation}
    y_l = \begin{cases}
        \text{Adapter}(x_{l-1} + x_l)+x_{l-1} & \text{if } l = 1, \\
        \text{Adapter}(y_{l-1} + x_l) + y_{l-1} & \text{otherwise}.
    \end{cases}
\end{equation}

The adapter used in SelaVPR consists solely of linear projections and a non-linearity activation, only operating on the channel dimension without modeling interactions along the spatial axes (i.e., between different patch tokens). This is still feasible for the common adaptation method of inserting the adapter inside the transformer block, because the MHA layer of subsequent blocks will model interactions between different tokens. However, it brings obvious limitations in the memory-efficient adaptation method, in which the output of the adapter will not be input again into the subsequent transformer blocks. So previous work \cite{metl} alternately applies the adapter along the channel- and token-dimensions to address it.

In this work, we introduce the multi-scale convolution (MultiConv) adapter, which is first proposed in our previous CricaVPR work \cite{cricavpr} but is used by inserting it into the transformer block (similar to SelaVPR but different from SelaVPR++). Notably, in CricaVPR, the MultiConv adapter is used to add multi-scale local prior knowledge to the model for enhancing the performance in the VPR task. However, in SelaVPR++, it not only introduces multi-scale local priors, but more importantly, it breaks the limitation that the vanilla adapter cannot model the interactions between different patch tokens. In the MultiConv adapter, we add a MultiConv module (with a skip connection) between the activation layer and the up-projection layer, as shown in Fig. \ref{MultiConv} (c). This module contains three parallel convolution paths of different scales (1$\times$1, 3$\times$3, 5$\times$5), which is inspired by the inception module in GoogLeNet \cite{inception}. The 1$\times$1 convolution is also applied before the 3$\times$3 and 5$\times$5 convolutions to reduce the number of channels. This design and the bottleneck structure make our MultiConv adapter still lightweight. The outputs of the three convolution paths are subsequently concatenated to achieve integration. The MultiConv adapter only processes patch tokens, which are reshaped into a $W\times H\times D$-dim (from $N\times D$) feature map to restore the spatial position before being input into the adapter. In this way, the spatially adjacent patch tokens can be modeled interactions via 3$\times$3 and 5$\times$5 convolutions. Moreover, since the MultiConv adapter has introduced multi-scale local prior knowledge to the model, we no longer need to purposefully design a separate local adaptation module to process local information for VPR as in our previous SelaVPR.

It is worth noting that both the work \cite{metl} and our work follow the memory-efficient adaptation paradigm proposed in previous work \cite{lst}. Apart from the slight differences in overall pipeline (e.g., the use of residual connections parallel to the added trainable modules, and the order in which features from backbone are refined), the key difference between the three lies in the design of the trainable modules: the work \cite{lst} uses lightweight transformer blocks; the work \cite{metl} uses token-channel alternating adapters; we use MultiConv adapters.

Additionally, it is worth mentioning that our design is plug-and-play and offers excellent scalability. We can flexibly apply varying numbers of MultiConv adapters for a trade-off between accuracy and efficiency. For instance, given a DINOv2 backbone based on ViT-L/14, consisting of a patch embedding layer and 24 transformer blocks, we generally equip it with 24 MultiConv adapters, one per block, as shown in Fig. \ref{MultiConv} (c). However, to further reduce memory usage and get higher efficiency, it is also feasible to apply adapters to only the last partial blocks, or only an adapter to every $m$ blocks. For example, if we apply a MultiConv adapter to every 3 blocks, it can be formulated as: 
\begin{equation}
    y_l = \begin{cases}
        \text{Adapter}(x_{l-1} + x_{3l}) + x_{l-1} & \text{if } l = 1, \\
        \text{Adapter}(y_{l-1} + x_{3l}) + y_{l-1} & 2 \leq l \leq 8.
    \end{cases}
\end{equation}
Furthermore, due to the intrinsic structure and parameters of the backbone remaining unchanged, we can easily create multiple independent adapter network branches alongside the backbone for different purposes, which will be further described in the following subsections.

\begin{figure*}
    \centering
    \includegraphics[width=0.95\linewidth]{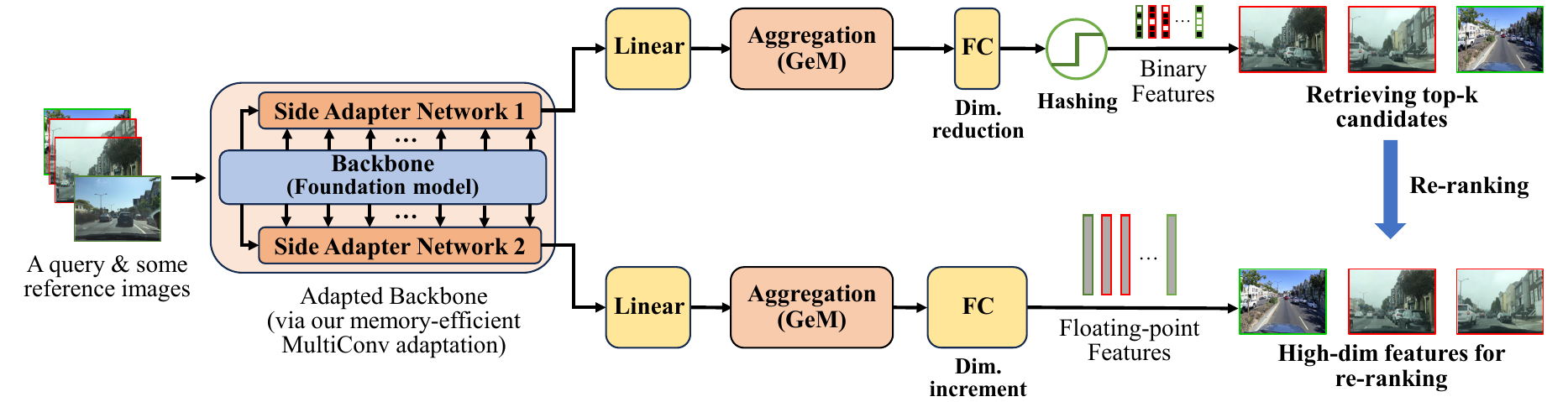}
    \vspace{-0.3cm}
    \caption
    {Illustration of our efficient two-stage VPR pipeline. The frozen foundation model combined with the side adapter networks is applied to extract the feature map. We leverage a linear projection and the GeM pooling (aggregation) to aggregate the feature map as a global descriptor. The branch above produces a compact binary feature for fast candidate retrieval. The branch below outputs a high-dimensional floating-point feature to re-rank the top-k candidates.}	
    \vspace{-0.3cm}
    \label{pipeline}
\end{figure*}

\subsection{An Efficient Two-Stage VPR Paradigm}
\label{sec:two-stage}
Existing two-stage VPR methods typically retrieve top-k candidate images using global features, and then re-rank these candidates by matching local features, which incurs significant overhead in computational latency, memory footprint, and storage usage (as mentioned in Section \ref{sec:2stagevpr}). To address this, we propose an innovative two-stage paradigm for more efficient VPR. Specifically, we use compact low-dimensional binary features for the initial retrieval to obtain top-k candidates, followed by employing robust high-dimensional floating-point features for re-ranking. In other words, our pipeline relies only on global features for both stages, without involving any local features. Moreover, compared to current one-stage methods, one key advantage of our pipeline is that the similarity of binary features is measured using the Hamming distance, which is more efficient than the Euclidean distance used for floating-point features, thereby speeding up the retrieval and providing candidate places quickly. Meanwhile, in contrast to the floating-point features of the same dimension, binary features require only 1/32 of storage. Besides, the binary features we use are low-dimensional, meaning that additional storage overhead is negligible.

Our pipeline is simple yet powerful, primarily composed of a pre-trained foundation model (i.e., DINOv2) as the backbone and two independent side adapter networks corresponding to two different branches respectively, as shown in Fig. \ref{pipeline}. The pre-trained backbone is completely frozen, while its two side adapter networks are trainable. The side network first uses cascaded MultiConv adapters to refine the intermediate features from the backbone, as described in the previous subsection. Subsequently, we apply a linear projection and an aggregation layer (i.e., GeM pooling \cite{gem}) to get the global place representation. After that, for the above binary feature branch, we use a fully-connected (FC) layer for dimensionality reduction, as well as a hashing operation to implement binary quantization, obtaining low-dimensional binary features (hash codes) to retrieve top-k candidates quickly. For the below floating-point feature branch, we leverage an FC layer to increase the dimension. Finally, we can get high-dimensional floating-point features to re-rank the candidates for better performance. Notably, our floating-point features used for re-ranking are higher-dimensional compared to binary features, but lower-dimensional than most SOTA methods \cite{salad,salad-cm,boq} due to the use of GeM pooling. In addition, we perform L2 normalization after the FC layer in both branches.

Since the two side adapter networks are independent of each other, our pipeline is flexible and scalable. To be specific, removing the binary feature branch makes our pipeline equivalent to current common one-stage VPR methods. In contrast, removing the other one enables very fast retrieval based on compact binary hash codes, which excels in limited-resource and large-scale VPR applications, particularly when high recognition performance is not urgently required. Additionally, we can either train both branches simultaneously, or first train only one branch and then use the obtained parameters to initialize the other branch for further training.

\subsection{Similarity-Constrained Deep Hashing}
Hashing has gained widespread attention in approximate nearest neighbor search for large-scale image retrieval due to its computational efficiency and retrieval quality by encoding data into binary vectors and performing search with Hamming distance. Deep hashing, in particular, has further enhanced performance by combining deep representation learning with hash coding. In this work, we attempt to adopt deep hashing to provide compact binary place representation, which is achieved by using a sign function $sgn(\cdot)$ to quantize floating-point features \{$\boldsymbol{f}_i$\} into binary hash codes \{$\boldsymbol{b}_i$\}. Formally,
\begin{equation}
    \boldsymbol{b}_i = sgn(\boldsymbol{f}_i).
\end{equation}
That is, negative values in the feature $\boldsymbol{f}_i$ are hashed to -1 in the feature $\boldsymbol{b}_i$, and all other values are +1. However, due to the gradient of hashing operation (i.e., sign function) being zero or infinite everywhere, how to achieve end-to-end training of the model is a pressing problem. A common method \cite{hashing2016} is to compute the metric loss (i.e., the loss used to push samples of different categories apart and pull samples of the same category together) using the features before quantization, and add a quantization loss (e.g., L1/L2 regulation loss) to reduce the quantization error between the floating-point features and hash codes (i.e., binary features). However, previous study \cite{lcdsh} has demonstrated that minimizing the L1/L2 quantization loss will conflict with the metric loss due to the contradictory learning purpose, and preserving the similarity of pair-wise features is more valuable than directly forcing the descriptors before and after hashing to be close. 

To better apply deep hashing to the place representation and optimize efficiency, we simplify previous work \cite{lcdsh} and propose a similarity-constrained loss $\mathcal{L}_Q$ to take the place of the common L1/L2 quantization loss. Concretely, given a pair-wise floating-point features \{$\boldsymbol{f}_i, \boldsymbol{f}_j$\} and a pair-wise quantized binary hash codes \{$\boldsymbol{b}_i, \boldsymbol{b}_j$\}, we need to minimize the loss
\begin{equation}
    \mathcal{L}_Q =\frac{\sum_{i,j} (s(\boldsymbol{f}_i, \boldsymbol{f}_j)  -  s(\boldsymbol{b}_i, \boldsymbol{b}_j))^2}{K},
    \label{eq:lq1}
\end{equation}
where $K$ represents the number of feature pairs (only using partial, e.g. 1/5, positive and negative pairs in a batch to reduce memory usage) and $s(\cdot , \cdot)$ denotes the cosine similarity, i.e.,
\begin{equation}
    s(\boldsymbol{x},\boldsymbol{y}) = \frac{\langle \boldsymbol{x}, \boldsymbol{y} \rangle}{\left\|\boldsymbol{x}\right\| \cdot \left\|\boldsymbol{y}\right\|}. 
\end{equation}
Since $\boldsymbol{f}_i$ and $\boldsymbol{f}_j$ are L2-normalized, the inner product is equivalent to the cosine similarity between them. For the hash code, the modulus $\left\|\boldsymbol{b}_i\right\|$ (or $\left\|\boldsymbol{b}_j\right\|$) is a constant $\sqrt{d}$, where $d$ is the dimension of features. So Eq. (\ref{eq:lq1}) can be rewritten as: 
\begin{equation}
    \mathcal{L}_Q = \frac{\sum_{i,j} (\langle\boldsymbol{f}_i, \boldsymbol{f}_j\rangle - \frac{\langle \boldsymbol{b}_i, \boldsymbol{b}_j \rangle}{d})^2}{K}.
\end{equation}

In contrast to \cite{lcdsh}, which applies the sigmoid transformation on inner products to reduce the sensitivity to large values in the features, our method directly preserves the cosine similarity of pair-wise features and can be simplified into a more concise form (without sigmoid). Due to the advanced backbone and metric loss to produce well-behaved features, as well as L2-norm on features in our work, we no longer consider the issue of harmful large values in features as in \cite{lcdsh}.

After obtaining $\mathcal{L}_Q(\boldsymbol{f},\boldsymbol{b})$, to train a deep hashing model, we can use the total loss $\mathcal{L}$ that combines the metric loss $\mathcal{L}_M(\boldsymbol{f})$ and similarity-constrained loss $\mathcal{L}_Q(\boldsymbol{f},\boldsymbol{b})$ by a weight $\lambda$ as: 
\begin{equation}
    \mathcal{L} = \mathcal{L}_M(\boldsymbol{f}) + \lambda \mathcal{L}_Q(\boldsymbol{f},\boldsymbol{b}).
    \label{eq:total_loss}
\end{equation}

Another solution to address the gradient issue of quantization operation is straight-through estimation (STE) \cite{ste}, which was proposed in the image compression area and addressed the gradient issue of rounding quantization function. Since the sign function in deep hashing is also a quantization function (which can be seen as a special form of rounding function), we employ it to perform the backpropagation of deep hashing in the VPR task. Specifically, we use the original sign function $sgn(\cdot)$ during forward propagation to obtain the binary hash codes. In back propagation, the derivative of binary hashing operation is replaced with the derivative of a smooth approximation $g$, which can be formulated as:  
\begin{equation}
    \frac{d}{d\boldsymbol{f}}sgn(\boldsymbol{f}) := \frac{d}{d\boldsymbol{f}}g(\boldsymbol{f}).
\end{equation}
Previous work \cite{ste} shows that using $g(x) = x$ is sufficient for a common quantization operation and here we also use it for the binary hashing operation. So
\begin{equation}
\frac{d\mathcal{L}_M(\boldsymbol{b})}{d\boldsymbol{f}} = \frac{d\mathcal{L}_M(\boldsymbol{b})}{d\boldsymbol{b}} \frac{d\boldsymbol{b}} {d\boldsymbol{f}}:= \frac{d\mathcal{L}_M(\boldsymbol{b})}{d\boldsymbol{b}} \frac{d} {d\boldsymbol{f}}g(\boldsymbol{f})=\frac{d\mathcal{L}_M(\boldsymbol{b})}{d\boldsymbol{b}}.
\end{equation}
In this way, the STE method achieves the integration of binary hashing operation into neural networks, and we can directly use the hashed binary descriptors to compute the metric loss $\mathcal{L}_M(\boldsymbol{b})$ for the end-to-end training.

In this work, we combine the above two methods, obtaining the final loss $\mathcal{L}$ by replacing the metric loss $\mathcal{L}_M(\boldsymbol{f})$ in Eq. (\ref{eq:total_loss}) with $\mathcal{L}_M(\boldsymbol{b})$. That is  
\begin{equation}
    \mathcal{L} = \mathcal{L}_M(\boldsymbol{b}) + \lambda \mathcal{L}_Q(\boldsymbol{f},\boldsymbol{b}). 
    \label{eq:final_loss}
\end{equation}

\subsection{Training Strategy and Unified VPR Training Dataset}
Our SelaVPR work \cite{selavpr} involves training the model on the Pitts30k \cite{pitts} and MSLS \cite{msls} datasets with a weakly supervised way using the triplet loss. However, some studies \cite{gsv,salad,boq} show that applying the multi-similarity (MS) loss \cite{multiloss} on large-scale curated datasets with full supervision can yield better performance. Thus, in this SelaVPR++ work, we train our model on the datasets with full supervision. Specifically, we follow the training framework of the GSV-Cities \cite{gsv} dataset using the MS loss as the metric loss for training, which can be expressed as follows:
\begin{equation}
\begin{split}
	\label{eq:MS}
	\mathcal{L}_{M} = \frac{1}{B}\sum_{q=1}^B  \bigg\{\frac{1}{\alpha}  { \log \big[1 + \sum_{p  \in \mathcal{P}_q } e^{-\alpha (S_{qp} - \gamma)}}\big]  \\
	+ \frac{1}{\beta }  { \log \big[1+ \sum_{n \in \mathcal{N}_q}
		 e^{\beta (S_{qn} - \gamma)} \big]} \bigg\},
\end{split}   
\end{equation}
where for each query $I_q$ in a batch, $\mathcal{P}_q$ and $\mathcal{N}_q$ are the sets of positive sample indices $\{p\}$ and negative sample indices $\{n\}$, respectively. $S_{qp}$ and $S_{qn}$ denote the cosine similarities of a positive pair $\{I_q,I_p\}$ and a negative pair $\{I_q,I_n\}$. $\alpha$, $\beta$, and $\gamma$ are three hyperparameters. 

Moreover, recent work SALAD-CM \cite{salad-cm} has significantly improved recognition performance on the VPR task by combining the GSV-Cities and MSLS datasets for training. In this paper, we attempt to merge several 
common training datasets (GSV-Cities, SF-XL \cite{cosplace}, Pitts30k, and MSLS) in a unified and simple protocol for robust VPR model training. It is worth noting that these datasets are annotated with different ways, thus working with different training protocols and loss functions. Specifically, for the SF-XL dataset, the CosPlace work \cite{cosplace}, which develops it, splits place images into a finite number of categories as GSV-Cities does. Therefore, we can directly merge (part of) SF-XL with GSV-Cities for training using the MS loss, although CosPlace uses a classification loss for training. However, for the Pitts30k and MSLS datasets, the place images are not split into limited categories, and previous works \cite{transvpr,selavpr} typically used triplet loss for weakly supervised training only with noisy GPS labels. In fact, the two datasets also provide the angle information. For Pitts30k, the images captured at the same place are from 2 pitch angles and 12 yaw angles, covering a diverse range of viewpoints. We create the pseudo angle labels for the 12 different yaw angles denoted as \{$0^\circ, 30^\circ, 60^\circ, \dots, 330^\circ$\}. For MSLS, we directly utilize the compass angles, which correspond to the absolute orientation. By combining the specific UTM coordinates (yielded by GPS) and angle information, we can follow the Cosplace work \cite{cosplace} to achieve place (category) division for the Pitts30k and MSLS datasets. Concretely, we merge the query images and database images of the training set, and split them geographically into square cells based on UTM coordinates \{\textit{east}, \textit{north}\}, then further divide each cell into a set of categories according to the orientation \{\textit{heading}\} of each image. Finally, the set of images divided into the category $C_{e_i,n_j,h_k}$ would be
\begin{equation}
    \{x:[\frac{east}{M}]=e_i,[\frac{north}{M}]=n_j,[\frac{heading}{\alpha}]=h_k\},
\end{equation}
where $M$ and $\alpha$ are hyperparameters, in meters and degrees respectively, to determine the range of each place category in position and angle. In our experiments, we set them to 10 meters and $60^\circ$ for SF-XL, but 15 meters and $60^\circ$ for Pitts30k and MSLS. Considering that similar images from the adjacent position (e.g., less than 1 meter) may be assigned different category labels due to this hard division, CosPlace trains the model using groups of non-adjacent categories rather than using all the categories at once. More specifically, this method yields $N\times N\times L$ groups by setting the minimum (translation and orientation) separation of two classes belonging to the same group, where $N$ and $L$ are two hyperparameters. In our experiments, we set them to 5 and 2 for SF-XL, but 3 and 2 for Pitts30k and MSLS. For SF-XL, we just leverage one group. For Pitts30k and MSLS, we use all groups. We use each group sequentially to avoid loading the place images, which are geographically adjacent but assigned to different labels, into the same batch for metric learning during training. 

In addition to SALAD-CM \cite{salad-cm} (combining GSV-Cities and MSLS for training with a complex mining strategy), the latest work SuperPlace \cite{superplace} also merges multiple datasets, i.e., GSV-Cities, SF-XL, MSLS, and Pitts250k (not Pitts30k as in ours), for model training and achieves excellent performance. Nevertheless, their processing methods for each dataset are different, hindering the integration of diverse datasets into a unified framework and even limiting the performance of trained models. For example, SuperPlace \cite{superplace} and our work both use the CosPlace \cite{cosplace} method to assign place images into a finite number of categories to process datasets other than GSV-Cities. However, SuperPlace fails to fully utilize the angle information of the Pittsburgh and MSLS datasets. It requires to implement local feature matching for the Pittsburgh dataset to assist in category division. For MSLS, it compromises by assuming that all images have the same orientation, without using compass angles data.
In summary, different from existing methods, we unify the framework of common training datasets with more consistent processing. This enables us to merge these datasets for better training of VPR models.

\section{Experiments}
\subsection{Datasets and Performance Evaluation}
\begin{table}
  \centering
  \caption{Summary of the Main Evaluation Datasets.}
  \vspace{-0.2cm}
  \label{tab:datasets}
  \begin{tabular}{cccc}
    \toprule
    \multirow{2}{*}{Dataset} & \multicolumn{1}{c}{\multirow{2}{*}{Description}} & \multicolumn{2}{c}{Number}  \\ \cline{3-4} 
     \multicolumn{1}{c}{}& & Database & Queries \\ 
     \midrule
    Pitts30k-test & urban, panorama & 10,000 & 6,816  \\ \hline
    MSLS-val &  urban, suburban & 18,871 & 740  \\
    MSLS-challenge & long-term & 38,770 & 27,092  \\ \hline
    Tokyo24/7 & urban, day/night & 75,984 &  315 \\ \hline
    Nordland & natural, seasonal & 27,592 &  27,592 \\
    \bottomrule
  \end{tabular}
  \vspace{-0.4cm}
\end{table}
Several VPR benchmark datasets mainly including Pitts30k \cite{pitts}, Tokyo24/7 \cite{densevlad}, MSLS \cite{msls}, and Nordland \cite{nordland} are used in our experiments. Table \ref{tab:datasets} summarizes their main information, and the details of them are as follows.

\textbf{Pitts30k} is collected from Google Street View panoramas, and provides 24 images with different viewpoints at each place. So this dataset shows large viewpoint changes. Pitts30k is a subset of Pitts250k (but harder than Pitts250k).

\textbf{Tokyo24/7} are captured from urban scenes and contains about 76k images in database. This dataset mainly exhibits viewpoint changes and drastic (day-night) condition changes.

\textbf{MSLS} is a large-scale VPR dataset containing over 1.6 million images labeled with GPS coordinates and compass angles, captured from 30 cities in urban, suburban, and natural scenes over 7 years. It covers various challenging visual changes due to illumination, weather, season, viewpoint, as well as dynamic objects, and includes subsets of training, public validation (MSLS-val), and withheld test (MSLS-challenge). The MSLS-val and MSLS-challenge sets are used to evaluate models.

\textbf{Nordland} primarily consists of suburban and natural place images, captured from the same viewpoint in the front of a train across four seasons, which allows the images to show severe condition (e.g., season) changes but no viewpoint variations. Following previous works \cite{benchmark,eigenplaces}, we use the winter images as queries and the summer images for reference.

We evaluate the recognition performance using Recall@N (R@N), which is the percentage of queries for which at least one of the N retrieved images is the right result. The threshold is set to 25 meters and 40$^\circ$ for MSLS, 25 meters for Pitts30k and Tokyo24/7, $\pm$10 frames for Nordland, following common evaluation procedures \cite{msls,benchmark}.

\subsection{Implementation Details}
We use DINOv2 (both the base and large versions) as the foundation model and conduct experiments on NVIDIA GeForce RTX 3090 GPUs using PyTorch. The resolution of the input image is 224$\times$224 in training and 322$\times$322 in inference. 
For the model using DINOv2-base, the token dimension in the backbone is 768 and the number of input channels for the MultiConv module is 384, that is, the bottleneck ratio of adapters is 0.5. The 1×1 convolution before the 3×3 and 5×5 convolution reduces the number of channels to 24. The numbers of output channels for the three convolutional paths (1×1, 3×3, and 5×5) are 192, 96, and 96, respectively. For the model using DINOv2-large, the token dimension in the backbone is 1024, and the number of channels for each convolution layer in the MultiConv adapter is also 1024/768 times higher than that for DINOv2-base. We set the hyperparameters of the MS loss as in \cite{gsv}. The coefficient $\lambda$ in Eq. (\ref{eq:final_loss}) is set to 0.1. We train models using the Adam optimizer with the initial learning rate set as 0.0004, halved every 3 epochs. A training batch consists of 120 places with 4 images each (i.e., 480 images). When the performance on MSLS-val does not have improvement for 12 epochs, the training is terminated. We also set the maximum epochs to 25. Additionally, for the DINOv2-base backbone, we equip each transformer block (12 blocks in total) with a MultiConv adapter. However, for the DINOv2-large backbone, we only equip the last 16 transformer encoder blocks with MultiConv adapters. This enables them to be trained on a single NVIDIA 3090 GPU with the above large batch size.

\begin{table*}[t]
    \centering
    \setlength{\tabcolsep}{0.24mm}
    \caption{Comparison to State-of-the-Art Methods on VPR Benchmark Datasets. The Best Is Highlighted in \textbf{Bold} and the Second Is \underline{Underlined}. $\dagger$ CricaVPR Uses a Cross-Image Encoder to Correlate Multiple Images from the Same Place to Achieve Better Performance on Pitts30k. It Is Not Included in the Comparison with Others on Pitts30k. The Results of BoQ Are Measured by Us Using the Official Code and Model Weights.}
    \vspace{-0.2cm}
    \begin{tabular}{@{}l|c|c|c||ccc||ccc||ccc||ccc||ccc||c}
    \toprule
    \multirow{2}{*}{Method} & \multirow{2}{*}{Dim} & \multirow{2}{*}{Backbone} &Training & \multicolumn{3}{c||}{Pitts30k-test} & \multicolumn{3}{c||}{Tokyo24/7} & \multicolumn{3}{c||}{MSLS-val} & \multicolumn{3}{c||}{MSLS-challenge} & \multicolumn{3}{c||}{Nordland} & Avg. \\
    \cline{5-20}
    & & &Dataset & R@1 & R@5 & R@10 & R@1 & R@5 & R@10 & R@1 & R@5 & R@10  & R@1 & R@5 & R@10 & R@1 & R@5 & R@10 & R@5\\
    \hline
    Patch-NetVLAD \cite{patchvlad} &/&VGG16&/ & 88.7 & 94.5 & 95.9 & 86.0 & 88.6 & 90.5 & 79.5 & 86.2 & 87.7 & 48.1 & 57.6 & 60.5 & 44.9 & 50.2 & 52.2 & 75.4 \\
    TransVPR \cite{transvpr} &/&/&/ &  89.0 &  94.9 &  96.2 & 79.0 &  82.2 & 85.1 & 86.8 & 91.2 & 92.4 & 63.9 & 74.0  & 77.5 & 63.5 & 68.5 & 70.2 & 82.2 \\
    SelaVPR \cite{selavpr} &/&DINOv2-L&/ & 92.8 & 96.8 & 97.7 & 94.0 & 96.8 & 97.5 & 90.8 & 96.4 & 97.2 & 73.5 & 87.5 & 90.6 & 87.3 & 93.8 & 95.6 & 94.3 \\
    \hline
    NetVLAD \cite{netvlad} & 32768 &VGG16& Pitts30k & 81.9 &91.2 &93.7 & 60.6 & 68.9 & 74.6 & 53.1 &66.5 &71.1 &35.1 & 47.4 & 51.7 & 6.4 & 10.1 & 12.5 & 56.8 \\
    SFRS \cite{sfrs} & 4096 &VGG16& Pitts30k & 89.4 & 94.7 & 95.9 & 81.0 & 88.3 & 92.4 & 69.2 & 80.3 & 83.1 & 41.6 & 52.0 & 56.3 & 16.1 & 23.9 & 28.4 & 67.8 \\
    CosPlace \cite{cosplace} &512 &VGG16& SF-XL & 88.4 & 94.5 & 95.7 & 81.9 & 90.2 & 92.7 & 82.8 & 89.7 & 92.0  & 61.4 & 72.0 & 76.6 & 58.5 & 73.7 & 79.4 & 84.0 \\
    MixVPR \cite{mixvpr} & 4096 &ResNet50& GSV & 91.5 & 95.5 & 96.3 & 85.1 & 91.7 & 94.3 & 88.0 & 92.7 & 94.6 & 64.0 & 75.9 & 80.6 & 76.2 & 86.9 & 90.3 & 88.5 \\
    EigenPlaces \cite{eigenplaces} &2048 &ResNet50& SF-XL &92.5 &96.8 &97.6 &93.0 &96.2 &97.5 &89.1 &93.8 &95.0  &67.4 &77.1 &81.7 & 71.2 & 83.8 & 88.1 & 89.5 \\
    CricaVPR$^\dagger$ \cite{cricavpr} & 4096 &DINOv2-B& GSV & 94.9$^\dagger$ & 97.3$^\dagger$ & 98.2$^\dagger$ & 93.0 & 97.5 & 98.1 & 90.0 &  95.4 & 96.4 & 69.0 & 82.1 & 85.7 & 90.7 & 96.3 & 97.6 & 93.7 \\
    SALAD \cite{salad} & 8448 &DINOv2-B& GSV & 92.5 & 96.4 & 97.5 & 94.6 & 97.5 & 97.8 & 92.2 & 96.4 & 97.0 & 75.0 & 88.8 & 91.3 & 89.7 & 95.5 & 97.0 & 94.9 \\
    SALAD-CM \cite{salad-cm}  & 8448 &DINOv2-B& GSV+MSLS & 92.7 & 96.8 & \underline{97.9} & 96.8 & 97.5 & 97.8 & \underline{94.2} & \underline{97.2} & \underline{97.4} & \underline{82.7} & 91.2 & 92.7 & \underline{96.0} & \underline{98.5} & \underline{99.2} & 96.2 \\
    BoQ \cite{boqArxiv} & 12288 &DINOv2-B& GSV & \underline{93.7} & \underline{97.1} & \underline{97.9} & 96.5 & \underline{97.8} & 98.4 & 93.8 & 96.8 & 97.0 & 79.0 & 90.3 & 92.0 & 90.6 & 96.0 & 97.5 & 95.6 \\ 
    \hline
    SelaVPR++(resource) & 2048 &DINOv2-B& Unified & 93.3 & 96.6 & 97.6 & \underline{97.5} & \textbf{98.7} & \underline{99.0} & \textbf{94.5} & 96.9 & 97.3 & 81.0 & \underline{91.9} & \underline{93.7} & 94.6 & 97.8 & 98.3 & \underline{96.4} \\ 
    SelaVPR++(performance)& 4096 &DINOv2-L & Unified & \textbf{94.4} & \textbf{97.5} & \textbf{98.1} & \textbf{98.1} & \textbf{98.7} & \textbf{99.4} & \textbf{94.5} & \textbf{98.0} & \textbf{98.2} & \textbf{84.0} & \textbf{93.7} & \textbf{94.4} & \textbf{97.2} & \textbf{99.0} & \textbf{99.4} &  \textbf{97.4} \\
    \bottomrule
    \end{tabular}
    \label{tab:compare_SOTA}
    \vspace{-0.2cm}
\end{table*}

\begin{table}
  \centering
  \setlength{\tabcolsep}{1.5mm}
  \caption{Comparison (R@1) to SOTA Methods on Supplementary Datasets. We Use the Performance-Focused SelaVPR++.}
  \vspace{-0.2cm}
  \begin{tabular}{l|ccccc}
  \toprule
    \multirow{2}{*}{Method} & \multirow{2}{*}{SPED} & \multirow{2}{*}{Eynsham} & SVOX & SVOX & SVOX 
    \vspace{-0.05cm} \\
    & & & -Night & -Rain & -Overcast \\
    \hline
    CosPlace \cite{cosplace} & 75.5 & 88.3 & 44.8 & 85.2 & 88.5 \\
    MixVPR \cite{mixvpr} & 85.2 & 89.4 & 64.4 & 91.5 & 96.2 \\
    EigenPlaces \cite{eigenplaces} & 70.2 & 90.7 & 58.9 & 90.0 & 93.1 \\
    SelaVPR \cite{selavpr} & 89.5 & 90.6 & 89.4 & 94.7 & 97.0 \\
    SALAD \cite{salad} & 92.1 & 91.6 & 95.4 & 98.5 & 98.3 \\
    SALAD-CM \cite{salad-cm} & 89.3 & 91.9 & 95.5 & 98.4 & \underline{98.5} \\
    BoQ \cite{boqArxiv} & \underline{92.5} & \underline{92.2} & \underline{97.7} & \underline{98.8} & \underline{98.5} \\
    \hline
    SelaVPR++ & \textbf{92.9} & \textbf{92.5} & \textbf{98.4} & \textbf{99.0} & \textbf{98.7} \\
    \bottomrule
    \end{tabular}
    \label{tab:compare_SOTA2}
    \vspace{-0.3cm}
\end{table}

\subsection{Comparison with State-of-the-Art Methods}
In this subsection, we compare our proposed SelaVPR++ with a wide range of SOTA VPR methods, including nine one-stage methods using global feature retrieval: NetVLAD \cite{netvlad}, SFRS \cite{sfrs}, CosPlace \cite{cosplace}, MixVPR \cite{mixvpr}, EigenPlaces \cite{eigenplaces}, CricaVPR \cite{cricavpr}, SALAD \cite{salad}, SALAD-CM \cite{salad-cm}, and BoQ \cite{boq}, as well as three two-stage methods with local feature re-ranking: Patch-NetVLAD \cite{patchvlad}, TransVPR \cite{transvpr}, and SelaVPR \cite{selavpr}. Note that the latest studies (i.e., SelaVPR, CricaVPR, SALAD, BoQ, and SALAD-CM) and our SelaVPR++ all use the foundation model DINOv2 for feature extraction and adopt different ways to fine-tune it for better performance. We provide two configurations of our SelaVPR++, i.e., a resource-focused configuration that uses the DINOv2-base backbone and outputs 2048-dim floating-point global descriptors for re-ranking, as well as a performance-focused configuration that uses the DINOv2-large backbone and outputs 4096-dim floating-point global descriptors for re-ranking. Both of them use 512-dim binary descriptors to retrieve top-100 candidate images. Besides, Cosplace and EigenPlaces build a large-scale dataset (SF-XL) for training. MixVPR, SALAD, CricaVPR, and BoQ all use GSV-Cities, whereas SALAD-CM combines both the GSV-Cities and MSLS-train datasets for training. Our approach unifies the Pitts30k-train, MSLS-train, SF-XL, and GSV-Cities datasets for more powerful model training.

\textbf{1) Quantitative results.} Table \ref{tab:compare_SOTA} shows the quantitative comparison results and our performance-focused SelaVPR++ achieves the best R@1/R@5/R@10 on all datasets. The methods using the foundation model DINOv2 as the backbone all achieve outstanding performance and surpass previous works (e.g., CosPlace, MixVPR, and EigenPlaces) on these datasets that exhibit real-world diverse challenges. This fully demonstrates that the foundation model can provide powerful feature representation with appropriate fine-tuning. Moreover, it is worth mentioning that, except for our SelaVPR++, no single method can consistently outperform others on all datasets. Specifically, BoQ and SALAD-CM, as two SOTA methods, exhibit their own distinct strengths. BoQ shows significant advantages over others on urban datasets, i.e., Pitts30k and Tokyo24/7. In contrast, SALAD-CM achieves better performance on the datasets covering suburban and natural scenes, i.e., MSLS and Nordland. However, our performance-focused SelaVPR++ gets the best results on all datasets, although the dimension of our global descriptors is less than half of theirs and we use more compact binary features to retrieve candidates. For example, although BoQ and SALAD-CM have achieved 93.7\% R@1 on Pitts30k and 96.0\% R@1 on Nordland respectively, our method can further improve R@1 to 94.4\% on Pitts30k and 97.2\% on Nordland. More importantly, our method achieves 93.7\% R@5 on MSLS-challenge and ranks first on the official leaderboard, which is a large improvement over the previous best result (91.2\% of SALAD-CM). Besides, our SelaVPR++ significantly outperforms its predecessor, our previous SelaVPR work, by 3.1\% for the average R@5 of five datasets. We also provide the comparison between our SelaVPR++ and other SOTA methods on some supplementary VPR datasets (i.e., SPED \cite{SPED}, Eynsham \cite{eynsham}, and SVOX \cite{SVOX}, all provided by \cite{benchmark}) as shown in Table \ref{tab:compare_SOTA2}, and SelaVPR++ still gets the best results. These results suggest that our method can produce highly robust global descriptors and excel in various challenging scenarios.

In addition, our resource-focused SelaVPR++ is also comprehensively superior to the vast majority of methods (e.g., EigenPlaces, CricaVPR, and SALAD) by a considerable margin, and outperforms BoQ and SALAD-CM for the average R@5. More importantly, compared to SALAD-CM (8448-dim) and BoQ (12288-dim), our resource-focused SelaVPR++ employs low-dimensional (2048-dim) global descriptors for re-ranking and more compact (512-dim) binary descriptors for initial retrieval. This indicates that our resource-focused model can achieve significantly higher retrieval efficiency, which will be further demonstrated in subsequent ablation experiments. It is worth mentioning that, even though our performance-focused SelaVPR++ uses a larger DINOv2-large backbone, its resource requirements are less than SALAD and BoQ (with the DINOv2-base backbone). We have measured the GPU memory usage of all three methods during training, and the results are summarized in Table \ref{tab: comparison with BoQ and SALAD in memory}. Our method consumes nearly half the memory compared to SALAD, and even less than BoQ, which only fine-tunes the last two blocks.

To ensure a fair comparison with BoQ and SALAD, as well as show the scalability of SelaVPR++, we additionally conduct experiments with more consistent settings, including the same backbone (DINOv2-base), training dataset, descriptor dimensionality, and image resolution (224$\times$224 in training, different from 280$\times$280 in the BoQ paper \cite{boqArxiv}). 
The first way to unify the dimensionality is to reduce the descriptor dimensionalities of BoQ and SALAD to the same as our method (2048-dim). This can be easily achieved by reducing the dimensionality of the linear layers in BoQ and the number of clusters in SALAD. The results are shown in the first three rows of Table \ref{tab: impact of our method on BoQ and SALAD}. SelaVPR++ outperforms SALAD (2048-dim) and BoQ (2048-dim) on all datasets. The second way to unify the dimensionality is to increase the descriptor dimensionality of SelaVPR++ to the same as SALAD and BoQ, respectively. Given that the default GeM pooling in SelaVPR++ is not suitable for high dimensionalities, we correspondingly replace GeM in SelaVPR++ with the BoQ/SALAD aggregator for comparison. Since our method is not in conflict with BoQ and SALAD, this is also easy to achieve.
The results are shown in the remaining rows of Table \ref{tab: impact of our method on BoQ and SALAD}, and we can summarize three points: \textbf{(1)} With the same training dataset, our method almost always outperforms BoQ/SALAD, which further highlights the strength of our method. \textbf{(2)} On the Pitts30k, MSLS-val, and Nordland datasets, both SelaVPR++(BoQ) and SelaVPR++(SALAD) trained on the unified dataset outperform SelaVPR++(resource) in Table \ref{tab:compare_SOTA}, showing that SelaVPR++ can be further improved with stronger aggregators, i.e., good scalability. \textbf{(3)} Other SOTA methods can also benefit a lot from our unified dataset, indicating that it is a universal contribution to VPR.

\begin{table}[t]
    \centering
    \setlength{\tabcolsep}{1.0mm}
    \caption{The Training GPU Memory Usage of Our Method, BoQ, and SALAD. We Consistently Set the Training Image Size to $224 \times 224$ and the Batch Size to 40 for All Methods.}
    \vspace{-0.2cm}
    \label{tab: comparison with BoQ and SALAD in memory}
    \begin{tabular}{l|c|c|c}
    \toprule
    Method & Backbone & Fine-tuning Technique & GPU Memory (GB)$ \downarrow $ \\
    \hline
    SelaVPR++ & DINOv2-L & Our adaptation method & \textbf{7.97} \\
    BoQ & DINOv2-B & Partial-tuning(2 blocks) & 8.20 \\
    SALAD & DINOv2-B & Partial-tuning(4 blocks) & 14.91 \\
    \bottomrule
    \end{tabular}  
    \vspace{-0.2cm}
\end{table}

\begin{table}[t]
    \centering
    \setlength{\tabcolsep}{0.2mm}
    \caption{Consistent Comparison to SALAD/BoQ. All Methods Use DINOv2-Base Backbone. SALAD (2048-Dim) and BoQ (2048-Dim) Are the Dimensionality-Reduced Versions of SALAD and BoQ, Which Are the Same 2048-Dim as Our SelaVPR++. SelaVPR++(SALAD) and SelaVPR++(BoQ) Use the SALAD and BoQ Aggregators to Replace GeM Pooling to Output Descriptors of the Same Dimensionality as SALAD and BoQ, Respectively, i.e., 8448-Dim and 12288-Dim. The Image Resolution Is $224\times224$ in Training and $322\times322$ in Inference.}
    \label{tab: impact of our method on BoQ and SALAD}
    \vspace{-0.2cm}
    \begin{tabular}{l|c|cc|cc|cc|cc}
    \toprule
    \multirow{2}{*}{Method} & Training & \multicolumn{2}{c|}{Pitts30k-test} & \multicolumn{2}{c|}{MSLS-val} & \multicolumn{2}{c|}{Tokyo24/7} & \multicolumn{2}{c}{Nordland} \\
    \cline{3-10}
    & Dataset & R@1 & R@5 & R@1 & R@5 & R@1 & R@5 & R@1 & R@5 \\
    \hline
    SALAD (2048-dim) & \multirow{3}{*}{GSV} & 91.5 & 96.1 & 91.2 & \textbf{96.2}  & 93.0 & 96.5 & 77.2 & 87.8 \\
    BoQ (2048-dim) & & 92.0 & 96.2 & 91.8 & 96.1 & 94.9 & 97.5 & 77.7 & 88.7 \\
    SelaVPR++ & & \textbf{92.6} & \textbf{96.5} & \textbf{92.6} & \textbf{96.2} & \textbf{96.2} & \textbf{97.8} & \textbf{81.3} & \textbf{91.3} \\
    \hline
    \hline
    SALAD & \multirow{4}{*}{GSV} & 92.5 & 96.4 & 92.2 & 96.4 & 94.6 & 97.5 & 89.7 & \textbf{95.5} \\ 
    SelaVPR++(SALAD) & & \textbf{93.2} &\textbf{96.9} & \textbf{93.5} & \textbf{97.0} & \textbf{95.6} & \textbf{97.8} & \textbf{89.9} & \textbf{95.5} \\
    \cline{3-10}
    BoQ & & 92.5 & 96.5 & 92.2 & \textbf{96.4} & \textbf{96.2} & \textbf{98.4} & 84.8 & 93.3 \\ 
    SelaVPR++(BoQ) & & \textbf{93.3} & \textbf{96.8} & \textbf{92.4} & \textbf{96.4} & \textbf{96.2} & 98.1 & \textbf{92.1} & \textbf{96.6} \\
    \hline
    SALAD & \multirow{4}{*}{Unified} & 93.0 & 96.6 & 94.5 & 97.2 & 95.6 & 97.8 & 94.6 & 98.2 \\
    SelaVPR++(SALAD) & & \textbf{93.8} & \textbf{97.3} & \textbf{94.6} & \textbf{97.6} & \textbf{96.5} & \textbf{98.1} & \textbf{96.7} & \textbf{98.9} \\  
    \cline{3-10}
    BoQ & & 93.0 & 96.5 & 94.8 & 97.4 & 96.2 & \textbf{98.4} & 96.5 & 98.8 \\ 
    SelaVPR++(BoQ) & & \textbf{94.1} & \textbf{97.1} & \textbf{94.9} & \textbf{97.6} & \textbf{96.8} & \textbf{98.4} & \textbf{97.6} & \textbf{99.2} \\ 
    \bottomrule
    \end{tabular}
    \vspace{-0.3cm}
\end{table}

\begin{figure*}[!htbp]
    \centering
    \includegraphics[width=0.9\linewidth]{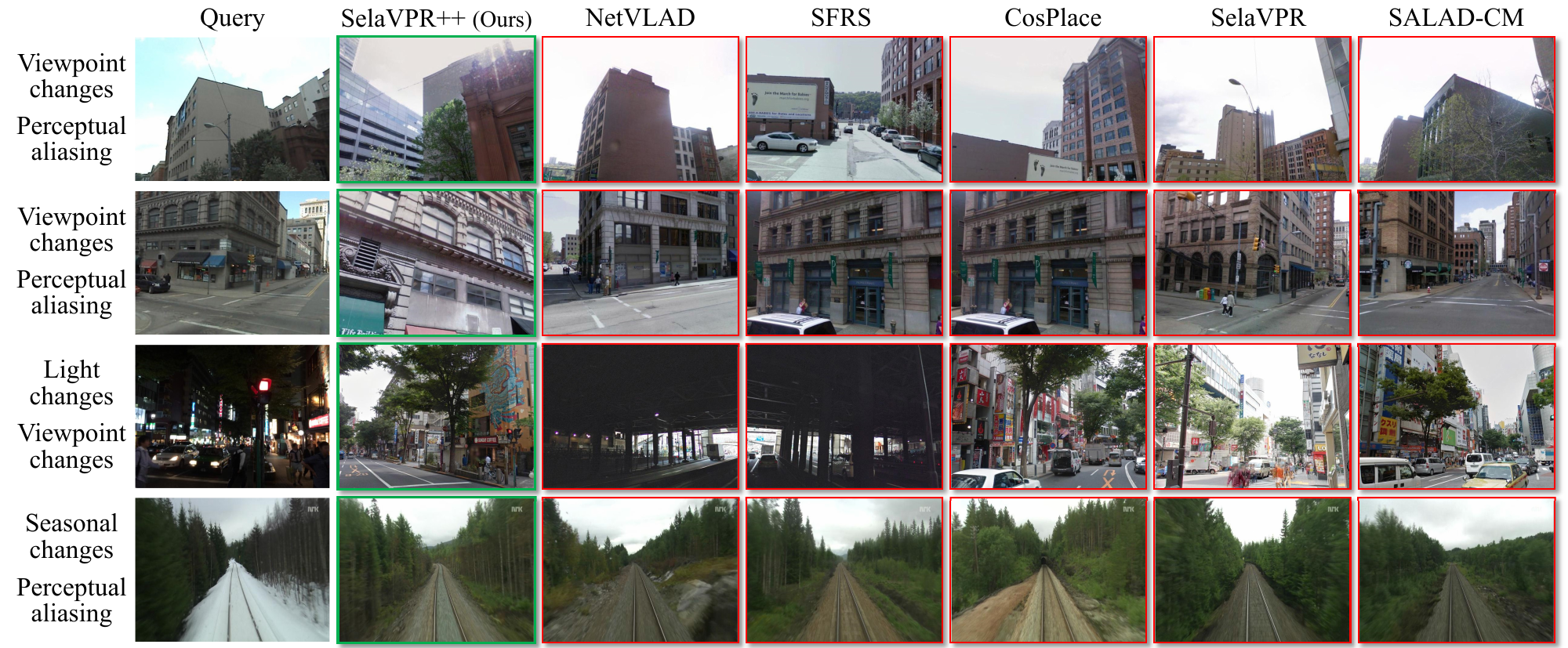}
    \vspace{-0.3cm}
    \caption{
    Qualitative results. In these challenging examples, our SelaVPR++ successfully returns the right database images, while other methods produce incorrect results. In the first two examples, which present drastic viewpoint changes between the query and (correct) database images, other methods wrongly return images similar to the query but from other places. The third example is quite challenging, as the query image is taken at night, showcasing significant changes in both light and viewpoint, with only the right part of the image recognizable but faintly (e.g., traffic light, coffee shop signboard, and special pattern on the building surface). The last query shows a natural scene and contains almost no landmarks. All of these examples require a powerful ability to capture discriminative place details and handle interference in order to obtain accurate results.
    }	
    \vspace{-0.3cm}
    \label{resultfig}
\end{figure*}

\begin{figure}[t]
    \centering
    \includegraphics[width=1.0\linewidth]{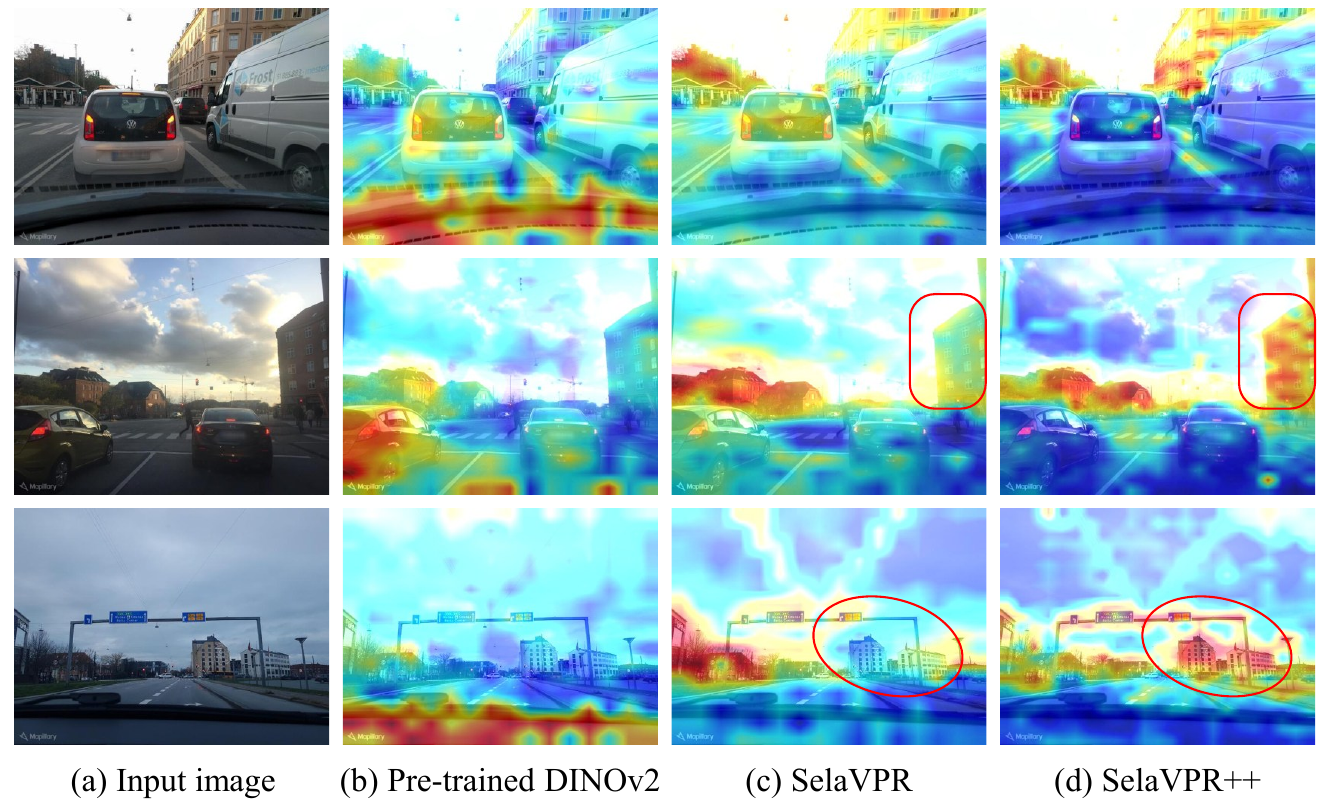}
    \vspace{-0.55cm}
    \caption{
    Heatmap visualizations of feature maps produced by the pre-trained foundation model (DINOv2), our previous work SelaVPR, and the proposed SelaVPR++. We compute the mean in the channel dimension of the output feature map and display it using the heatmap. SelaVPR aims to seamlessly adapt pre-trained models to VPR. However, it does not do well in detail for these examples. In the first two examples, it is still disturbed by a few dynamic vehicles. Besides, in the last two examples, SelaVPR does not pay enough attention to the partial discriminative buildings in the right region of the image or even ignores them. In contrast, SelaVPR++ can significantly filter out the dynamic foreground and focus on static landmarks useful for VPR.
    }	
    \vspace{-0.6cm}
    \label{attention-map}
\end{figure}

\textbf{2) Qualitative results.} Fig. \ref{resultfig} presents the top-1 database images retrieved by different methods for four query examples, which qualitatively demonstrates that our approach is highly robust against diverse challenges in VPR. These challenges mainly contain drastic viewpoint and condition (e.g., light and season) changes. In addition, there exist some images in the database that are visually similar to the query image but actually from different places, prone to result in perceptual aliasing. In almost every example, there are other methods that retrieve similar but wrong results due to perceptual aliasing, and our previous SelaVPR work is no exception. On the contrary, our SelaVPR++ can address all of these issues and return the correct images, demonstrating the excellent ability to capture discriminative place details and handle interference.

Moreover, to more intuitively illustrate that our SelaVPR++ can perform well in the VPR task, we visualize the attention map of the output feature map (before aggregation) from different models, including the pre-trained foundation model DINOv2, our previous SelaVPR work and the proposed SelaVPR++. As shown in Fig. \ref{attention-map}, the pre-trained model tends to pay attention to dynamic foreground objects (e.g., pedestrians and vehicles), and often ignores some static discriminative backgrounds (e.g., buildings and vegetation) that are useful to identify places. SelaVPR ameliorates the disadvantage to some extent through global adaptation and focuses more on static buildings. However, SelaVPR still occasionally pays some attention to dynamic vehicles or leaves out useful buildings in some challenging cases. SelaVPR++ can perform better than SelaVPR in detail and obviously distinguish static discriminative landmarks from the dynamic foreground, which is important for a robust VPR model. This demonstrates that SelaVPR++ can better bridge the gap between the model pre-training and VPR tasks compared to SelaVPR.

\subsection{Ablation Study}
In this subsection, we conduct a series of ablation experiments to demonstrate the effectiveness of several key components in SelaVPR++. We use floating-point features for direct retrieval by default, and set up two separate experiments to explore the two-stage VPR paradigm and deep hashing.

\textbf{1) Effect of memory-efficient MultiConv adaptation.}
To validate the advantages of our Memory-Efficient MultiConv Adaptation (abbreviated as MemEfficient-MCA), we compare it with the following five model fine-tuning methods:
\begin{itemize}
\item\underline{Freezing}: The backbone is frozen (used in AnyLoc \cite{anyloc}).
\item\underline{Full-tuning}: We fully fine-tune the model as a reference.
\item\underline{Partial-tuning}: We fine-tune the last partial layers (last 4 transformer blocks) of the backbone as in SALAD \cite{salad}.
\item\underline{SelaVPR-global}: We follow the global adaptation method in SelaVPR \cite{selavpr} (without local re-ranking), which is an adapter-based parameter-efficient fine-tuning method.
\item\underline{MemEfficient-vanilla}: We use the memory-efficient adaptation method with the vanilla adapter (i.e., replace MultiConv adapters with vanilla adapters) as another reference.
\end{itemize}
It is worth noting that the above fine-tuning methods are only applied to the backbone, while the aggregator (the GeM layer and the two linear layers before and after it) is always trainable. We use both DINOv2-base and DINOv2-large for experiments, outputting 2048-dim and 4096-dim global features for retrieval respectively. The GSV-Cities dataset is used for training. Since the Full-tuning will take up a huge amount of GPU memory, we adjust the batch size to 60 for all models based on DINOv2-large. Even so, the Full-tuning method still requires 6 NVIDIA 3090 GPUs. The results are shown in Table \ref{tab:ablation_adapterB} (DINOv2-base) and Table \ref{tab:ablation_adapterL} (DINOv2-large).
We summarize the two tables in the following points: 
\textbf{(1)} For the Freezing method, since our aggregator contains two trainable linear layers that can also adapt the features produced by backbone to the VPR task, it achieves decent results for partial datasets. However, for the Nordland dataset, which is mainly collected in natural scenes and across seasons, the model needs to be able to capture detailed information in the image that can distinguish places. Simply adjusting the feature map output by the frozen backbone cannot effectively bridge the gap between the tasks of model pre-training and VPR. Therefore, the performance of this method is much lower than other methods. \textbf{(2)} As for Full-tuning, when using the DINOv2-base backbone, the difference between it and other methods is not large on most datasets. However, it is obviously inferior to other methods on Tokyo247, which contains many nighttime images. It is because there is a generalization gap between Tokyo24/7 and the GSV-Cities training dataset (without night images), and full fine-tuning damages the excellent representation ability of the pre-trained foundation model (i.e., catastrophic forgetting). This phenomenon is more obvious for DINOv2-large that has a larger number of parameters. \textbf{(3)} The performance of the PEFT methods (i.e., SelaVPR-global, MemEfficient-vanilla, and MemEfficient-MCA) when using DINOv2-large is generally better than that of DINOv2-base. However, the R@1 results of the Full-tuning and Partial-tuning methods on Tokyo247 decrease when using DINOv2-large. Especially for Full-tuning, the R@1 on Pitts30k and MSLS-val also decreases. This shows that the PEFT method is more suitable for large models than direct (full or partial) fine-tuning. \textbf{(4)} Both the Partial-tuning and PEFT methods achieve good performance on all these datasets. However, compared with the previously proven Partial-tuning and SelaVPR, the memory-efficient adaptation method used in this paper can achieve better results. Furthermore, our MemEfficent-MCA upgrades vanilla adapters to MultiConv adapters, which can facilitate feature interactions along the spatial axes and introduce the multi-scale local priors. Thus, it achieves the best performance among these methods.

\begin{table}[t]
    \centering
    \setlength{\tabcolsep}{0.9mm}
    \caption{The Results of Different Fine-Tuning Methods with the DINOv2-Base Backbone. Models Are Fine-Tuned on GSV-Cities and Batch Size Is Set to 120.}
    \vspace{-0.2cm}
    \begin{tabular}{l|cc|cc|cc|cc}
    \toprule
    \multirow{2}{*}{Method} & \multicolumn{2}{c|}{Pitts30k-test} & \multicolumn{2}{c|}{MSLS-val} & \multicolumn{2}{c|}{Tokyo24/7} & \multicolumn{2}{c}{Nordland} \\
    \cline{2-9}
    & R@1 & R@5 & R@1 & R@5 & R@1 & R@5 & R@1 & R@5 \\
    \hline
    Freezing & 91.0 & 96.1 & 87.2 & 94.1 & 94.0 & 96.8 & 53.2 & 68.6 \\
    Full-tuning & 91.5 & 95.4 & 89.6 & 95.3 & 90.2 & 95.2 & 73.7 & 86.0 \\
    Partial-tuning & 92.0 & 96.3 & 90.0 & 96.1 & 94.9 & 97.1 & 77.4 & 88.5 \\
    SelaVPR-global & 92.0 & 96.1 & 90.4 & 95.3 & 93.0 & 96.5 & 73.3 & 85.2 \\
    \hline
    MemEfficient-vanilla & 91.9 & 96.1 & 91.6 & 96.1 & 95.9 & 97.1 & 75.0 & 86.8 \\
    MemEfficient-MCA & \textbf{92.6} & \textbf{96.5} & \textbf{92.6} & \textbf{96.2} & \textbf{96.2} & \textbf{97.8} & \textbf{81.3} & \textbf{91.3} \\
    \bottomrule
    \end{tabular} 
    \vspace{-0.2cm}
    \label{tab:ablation_adapterB}
\end{table}

\begin{table}[t]
    \centering
    \caption{The Results of Different Fine-Tuning Methods with the DINOv2-Large Backbone. Models Are Fine-Tuned on GSV-Cities and Batch Size Is Set to 60.}
    \vspace{-0.2cm}
    \setlength{\tabcolsep}{0.9mm}{
    \begin{tabular}{l|cc|cc|cc|cc}
    \toprule
    \multirow{2}{*}{Method} & \multicolumn{2}{c|}{Pitts30k-test} & \multicolumn{2}{c|}{MSLS-val} & \multicolumn{2}{c|}{Tokyo24/7} & \multicolumn{2}{c}{Nordland} \\
    \cline{2-9}
    & R@1 & R@5 & R@1 & R@5 & R@1 & R@5 & R@1 & R@5  \\
    \hline
    Freezing & 92.2 & 96.6 & 90.3 & 95.7 & 95.2 & \textbf{98.4} & 57.8 & 72.6 \\
    Full-tuning & 91.3 & 95.8 & 89.3 & 94.7 & 88.9 & 94.6 & 75.8 & 87.1 \\
    Partial-tuning & 92.6 & 96.4 & 90.4 & 96.2 & 93.7 & 97.1 & 78.8 & 89.0 \\
    SelaVPR-global & 92.5 & 96.7 & 90.4 & 95.8 & 94.6 & 96.8 & 75.7 & 86.4 \\
    \hline
    MemEfficient-vanilla & 92.5 & 96.8 & 92.7 & 96.5 & \textbf{96.5} & 97.8 & 79.9 & 90.2 \\
    MemEfficient-MCA & \textbf{93.3} & \textbf{97.2} & \textbf{93.4} & \textbf{96.8} & 96.2 & \textbf{98.4} & \textbf{81.4} & \textbf{91.1} \\
    \bottomrule
    \end{tabular}} 
    \label{tab:ablation_adapterL}
    \vspace{-0.2cm}
\end{table}

\begin{table}[t]
    \centering
    \setlength{\tabcolsep}{0.4mm}
    \caption{Comparison between the Memory-Efficient Adaptation Methods with the Vanilla Adapter, the Token-Channel Alternating Adapter, and Our MultiConv Adapter. We Use the DINOv2-Base Backbone, the GeM Aggregator (2048-Dim Output), and GSV-Cities for Training. The Resolution of the Input Image Is $224\times224$ in Both Training and Inference.}
    \vspace{-0.2cm}
    \label{tab: token-channel alternating adapter}
    \begin{tabular}{l|cc|cc|cc|cc}
    \toprule
    \multirow{2}{*}{Method} & \multicolumn{2}{c|}{Pitts30k-test} & \multicolumn{2}{c|}{MSLS-val} & \multicolumn{2}{c|}{Tokyo24/7} & \multicolumn{2}{c}{Nordland} \\
    \cline{2-9}
    & R@1 & R@5 & R@1 & R@5  & R@1 & R@5 & R@1 & R@5 \\
    \hline
    MemEfficient-vanilla  & 91.2 & 95.7 & 90.3 & 95.6 & 93.7 & 96.5 & 69.2 & 83.2 \\
    MemEfficient-alternating & 91.6 & 95.8 & 90.8 & \textbf{95.7} & 93.7 & 97.1 & 73.1 & 86.7 \\
    MemEfficient-MCA & \textbf{91.8} & \textbf{96.2} & \textbf{91.2} & \textbf{95.7} & \textbf{94.0} & \textbf{97.5} & \textbf{75.2} & \textbf{87.9} \\
    \bottomrule
    \end{tabular}
    \vspace{-0.2cm}
\end{table}

\begin{figure}[t]
    \centering
    \includegraphics[width=1.0\linewidth]{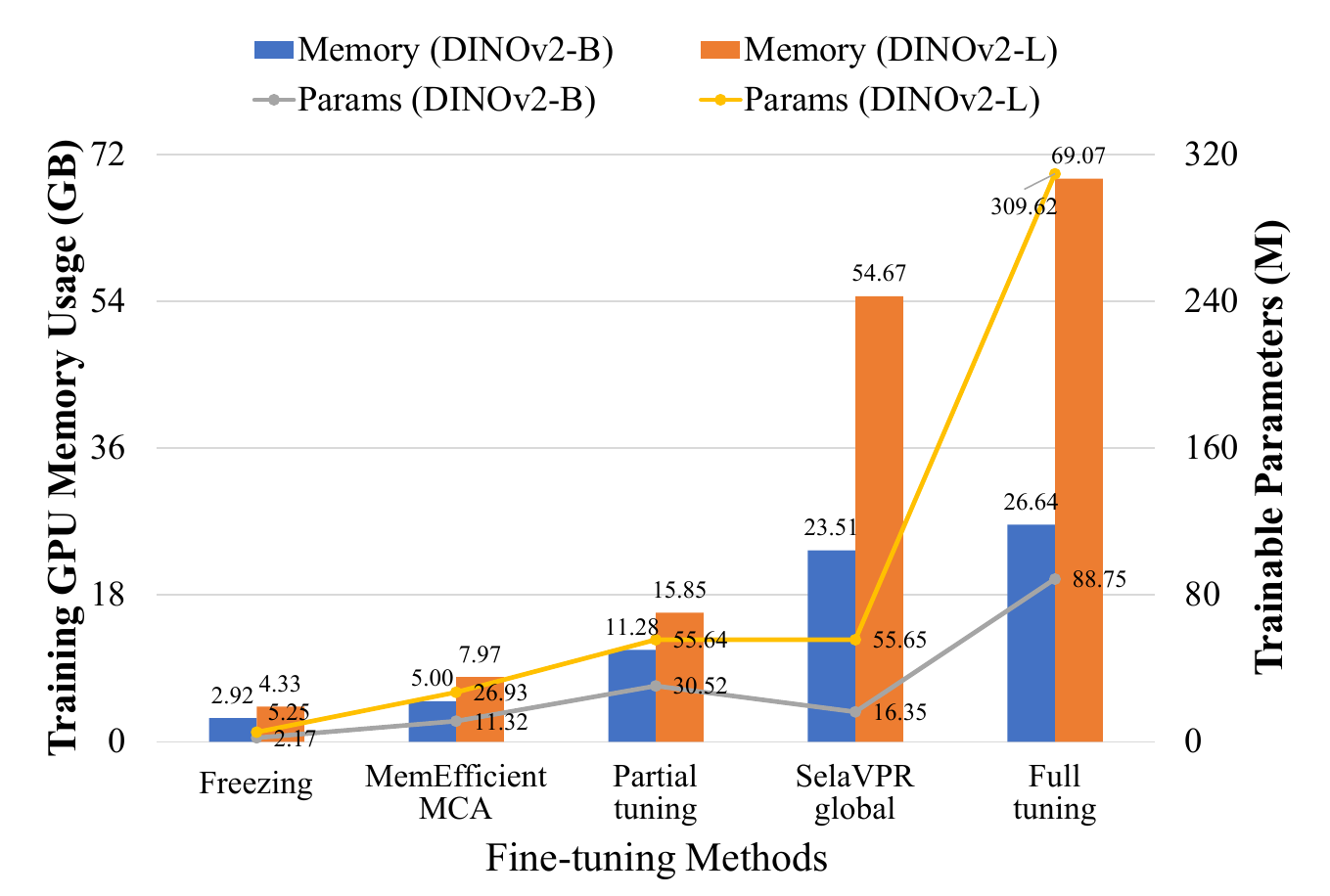}
    \vspace{-0.7cm}
    \caption{
    The comparison of different fine-tuning methods in training GPU memory usage (GB) and the number of trainable parameters (M). Due to the significant GPU memory requirements of Full-tuning, we consistently use an NVIDIA A800 with 80GB of memory capacity to measure the memory usage during training. Despite this, we still need to reduce the batch size to 40.
    }
    \vspace{-0.3cm}
    \label{fig:memory-params}
\end{figure}

\begin{table}[t]
    \centering
    \caption{The Comparison of Different Fine-Tuning Methods in Training Time for a Single Epoch on GSV-Cities. MemEfficient-MCA(4) Indicates Only Applying 4 Adapters for the Last 4 Blocks. We Consistently Train the Models on an NVIDIA A800 GPU with Batch Size Set to 40.}
    \vspace{-0.2cm}
    \setlength{\tabcolsep}{1.5mm}
    \begin{tabular}{l|c|c}
    \toprule
    Method & Backbone & Training Time (min/epoch) \\
    \hline
    Freezing & \multirow{5}{*}{DINOv2-B} & 6.75 \\
    MemEfficient-MCA && 6.85 \\
    Partial-tuning && 6.95 \\
    SelaVPR-global && 8.28 \\
    Full-tuning && 8.87 \\ 
    \hline
    Freezing & \multirow{6}{*}{DINOv2-L} & 8.80 \\
    MemEfficient-MCA && 11.97 \\
    MemEfficient-MCA(4) && 9.60 \\
    Partial-tuning && 11.47  \\
    SelaVPR-global && 23.12 \\
    Full-tuning && 24.73 \\ 
    \bottomrule
    \end{tabular}
    \label{tab:ablation_trainingtime}
    \vspace{-0.2cm}
\end{table}

Moreover, we also compare our method with the memory-efficient adaptation \cite{metl} that alternately applies adapters along the channel- and token-dimensions, denoted as MemEfficient-alternating. Due to the inherent limitation of the token-channel alternating adapter, it lacks adaptability to different image resolutions. Consequently, it can only be evaluated on images with the same resolution as in training. We compared three memory-efficient adaptation methods by setting the resolution to 224$\times$224 in both training and inference. The result is shown in Table \ref{tab: token-channel alternating adapter}. Compared to MemEfficient-vanilla, MemEfficient-alternating enables patch token interactions, resulting in better results. Furthermore, our MultiConv adapter not only models spatial interactions between patch tokens but also introduces multiscale local priors beneficial for VPR, thus achieving the best results. More importantly, both MemEfficient-vanilla and ours can improve performance by increasing the resolution in inference (better results in Table \ref{tab:ablation_adapterB} using 322$\times$322 resolution in inference). However, this cannot be achieved for MemEfficient-alternating.

In addition to the recognition performance, we also evaluate the number of trainable parameters and GPU memory usage, with the results shown in Fig. \ref{fig:memory-params}. Since the aggregator needs to be trained, the Freezing method also occupies a certain amount of GPU memory during training.
For SelaVPR-global, the number of trainable parameters is less than 1/5 of Full-tuning, and it is also not inferior to Partial-tuning. However, its memory usage is more than 3/4 of that of Full-tuning, and much higher than that of Partial-tuning. So the adapter-based PEFT method used by SelaVPR-global is efficient in parameter, but not efficient in memory. For our MemEfficient-MCA, its trainable parameters are less than half of those in Partial-tuning for both DINOv2-large and DINOv2-base, and it also only takes up less than half of GPU memory used by Partial-tuning for DINOv2-base (50.3\% for DINOv2-large). Furthermore, for DINOv2-large, the GPU memory occupied by our method is only 14.6\% of that used by SelaVPR-global and 11.5\% of Full-tuning. From these comparisons, it is clear that our method is efficient in both parameter and memory.

Meanwhile, since the gradient backpropagation in our method no longer passes through the backbone during training, it saves a lot of training time. Table \ref{tab:ablation_trainingtime} shows the training time of different fine-tuning methods for training an epoch on GSV-Cities (using an NVIDIA A800). Compared with Full-tuning, SelaVPR-global does not save significant time, that is, it is also not efficient in training time. For DINOv2-base, our MemEfficient-MCA consumes only 6.85 min/epoch, which is less than Partial-tuning (both are only slightly more than Freezing). For DINOv2-large, our method applies adapters to the last 16 blocks (DINOv2-base has only 12 blocks), while Partial-tuning still only fine-tunes the last 4 blocks. Therefore, the latter achieves a slightly shorter training time. However, our approach can further improve efficiency by reducing the number of adapters. When we use only 4 adapters, i.e., MemEfficient-MCA(4), the training time is 16.3\% less than Partial-tuning. These results show that our method is efficient in training time in addition to parameter and memory. In the next ablation experiment, we will demonstrate that our method can still achieve excellent recognition performance when using only a small number of adapters.

\begin{table}[t]
    \centering
    \setlength{\tabcolsep}{0.6mm}
    \caption{The Results of Using Different Numbers of Adapters (with Two Ways). All Models Are Based on DINOv2-Large Backbone and Trained on Our Unified Dataset (Batch Size is 120).}
    \vspace{-0.2cm}
    \begin{tabular}{l||cc||cc||cc||cc||cc}
    \toprule
    \multirow{2}{*}{Method} & \multicolumn{2}{c||}{Pitts30k-test} & \multicolumn{2}{c||}{MSLS-val} & \multicolumn{2}{c||}{Tokyo24/7} & \multicolumn{2}{c||}{Nordland} & \multicolumn{2}{c}{Average} \\
    \cline{2-11}
    & R@1 & R@5 & R@1 & R@5 & R@1 & R@5 & R@1 & R@5 & R@1 & R@5 \\
    \hline
    Freezing & 92.7 & 96.8 & 92.0 & 96.8 & 95.6 & 98.4 & 69.7 & 82.8 & 87.5 & 93.7 \\
    All & 94.3 & 97.4 & 94.7 & 97.7 & 97.5 & 98.7 & 96.2 & 98.8 & 95.7 & 98.2 \\ 
    Every 2 & 94.3 & 97.2 & 94.6 & 97.3 & 97.1 & \textbf{99.7} & 96.5 & 98.9 & 95.6 & 98.3 \\
    Every 3 & \textbf{94.4} & \textbf{97.5} & 94.2 & 97.3 & 97.5 & 98.7 & 96.4 & 98.9 & 95.6 & 98.1 \\
    Every 6 & 93.0 & 96.8 & 93.8 & 96.9 & 97.5 & 98.4 & 93.7 & 97.8 & 94.5 & 97.6 \\
    Last 16 & \textbf{94.4} & \textbf{97.5} & 94.3 & \textbf{98.0} & \textbf{98.1} & 99.0 & \textbf{97.0} & \textbf{99.1} & \textbf{96.0} & \textbf{98.4} \\
    Last 12 & 94.3 & \textbf{97.5} & 94.7 & 97.7 & 97.8 & 98.7 & 96.6 & 98.8 & 95.9 & 98.2 \\
    Last 8 & 94.0 & 96.8 & \textbf{94.9} & 97.6 & 97.1 & \textbf{99.7} & 95.5 & 98.6 & 95.4 & 98.2 \\
    Last 4 & 94.0 & 97.3 & 94.1 & 97.2 & 97.5 & 99.4 & 92.5 & 97.2 & 94.5 & 97.8 \\
    \bottomrule
    \end{tabular}  
    \vspace{-0.3cm}
    \label{tab:ablation_number_adapter}
\end{table}

\textbf{2) Effect of the number of adapters.}
As described in Section \ref{sec:adaptation}, our approach can further improve efficiency by reducing the number of adapters in two ways, namely, applying one adapter per $m$ blocks or applying adapters only to the last partial blocks. We equip DINOv2-large with different numbers of adapters and train them on our unified dataset, yielding the results shown in Table \ref{tab:ablation_number_adapter}. All methods equipped with adapters achieve good results and are significantly better than the baseline (i.e., Freezing). We mainly observe the performance of each setting through the average of R@N, which shows that the more adapters used in both ways, the better the performance. An exception is that the performance of applying adapters to the last 16 blocks is better than that for all blocks, indicating that the former has basically reached saturation performance. Therefore, for DINOv2-large, our recommendation is to only apply adapters to the last 16 blocks. In addition, the number of adapters used when applying an adapter per 2, 3, and 6 blocks just corresponds to applying adapters to the last 12, 8, and 4 blocks, and the results of the two are very close. So we believe both ways are good choices. Finally, reducing the number of adapters for the sake of efficiency is also advisable, because our method with only 4 adapters also achieves promising performance.

\begin{table*}[t]
    \centering
    \setlength{\tabcolsep}{0.8mm}
    \caption{The Results of Direct Retrieval Using Binary Features and Floating-Point Features, as Well as Our Re-Ranking Method. The R@100 Indicates the Upper Limit of (R@1/R@5/R@10) Performance After Re-Ranking, and Is Not Included in the Comparison.}
    \vspace{-0.2cm}
    \begin{tabular}{l|c||cccc||cccc||cccc||cccc}
    \toprule
    \multirow{2}{*}{Method} &  \multirow{2}{*}{Backbone} &\multicolumn{4}{c||}{Pitts30k-test} & \multicolumn{4}{c||}{MSLS-val} & \multicolumn{4}{c||}{Tokyo24/7} & \multicolumn{4}{c}{Nordland} \\
    \cline{3-18}
    & & R@1 & R@5 & R@10 & R@100 & R@1 & R@5 & R@10 & R@100 & R@1 & R@5 & R@10 & R@100 & R@1 & R@5 & R@10 & R@100 \\
    \hline
    Binary (512D) & \multirow{3}{*}{DINOv2-B} & 89.0 & 95.2 & 96.7 & (98.9) & 90.1 & 95.7 & 96.5 & (98.5) & 90.5 & 96.2 & 98.1 & (99.4) & 78.4 & 90.8 & 94.0 & (99.1) \\
    Float (2048D) && \textbf{93.3} & \textbf{96.6} & 97.4 & (99.5) & 94.3 & \textbf{96.9} & \textbf{97.6} & (99.1) & 96.8 & 98.4 & 98.7 & (99.7) & \textbf{94.7} & \textbf{98.3} & \textbf{99.0} & (99.9) \\
    Our re-ranking && \textbf{93.3} & \textbf{96.6} & \textbf{97.6} & (98.9) & \textbf{94.5} & \textbf{96.9} & 97.3 & (98.5) & \textbf{97.5} & \textbf{98.7} & \textbf{99.0} & (99.4) & 94.6 & 97.8 & 98.3 & (99.1) \\
    \hline
    Binary (512D) & \multirow{3}{*}{DINOv2-L} & 91.6 & 96.3 & 97.4 & (99.3) & 91.2 & 96.8 & 97.4 & (98.8) & 93.0 & 96.8 & 97.5 & (100.0) & 87.8 & 96.0 & 97.6 & (99.8) \\
    Float (4096D) && \textbf{94.4} & \textbf{97.5} & \textbf{98.1} & (99.4) & 94.3 & \textbf{98.0} & \textbf{98.2} & (99.2) & \textbf{98.1} & \textbf{99.0} & \textbf{99.7} & (99.7) & 97.0 & \textbf{99.1} & \textbf{99.5} & (100.0) \\
    Our re-ranking && \textbf{94.4} & \textbf{97.5} & \textbf{98.1} & (99.3) & \textbf{94.5} & \textbf{98.0} & \textbf{98.2} & (98.8) & \textbf{98.1} & 98.7 & 99.4 & (100.0) & \textbf{97.2} & 99.0 & 99.4 & (99.8) \\
    \bottomrule
    \end{tabular} 
    \vspace{-0.3cm}
    \label{tab:ablation_rerankparadigm}
\end{table*}

\begin{table}[t]
    \centering
    \setlength{\tabcolsep}{0.8mm}
    \caption{The Retrieval Latency Comparison of Different Methods for a Single Query on the Pitts30k Dataset. The Initial Retrieval (to Get Candidates or Only One-Stage) is Denoted as ``Initial".}
    \vspace{-0.2cm}
    \begin{tabular}{l|c|c|c|c}
    \toprule
    \multirow{2}{*}{Method} & Distance & \multicolumn{1}{c|}{Initial} & \multicolumn{1}{c|}{Re-ranking} & \multicolumn{1}{c}{Total Time} \\
    & Measurement & (ms) & (ms) & (ms) \\
    \hline
    TransVPR \cite{transvpr} & L2 & 1.94 & 3096.66 & 3098.60 \\
    SelaVPR \cite{selavpr} & L2 & 8.02 & 67.51 & 75.53 \\
    \hline
    Float (4096D) & L2 & 32.22 & / & 32.22 \\
    Float (2048D) & L2 & 16.00 & / & 16.00 \\
    Float (512D) & L2 & 3.89 & / & 3.89 \\
    Binary (512D) & Hamming &  0.26 & / & 0.26 \\
    \hline
    SelaVPR++ (resource) & Hamming+L2 & 0.26 & 0.14 & 0.40 \\
    SelaVPR++ (performance) & Hamming+L2 & 0.26 & 0.25 & 0.51 \\
    \bottomrule
    \end{tabular}
    \vspace{-0.2cm}
    \label{tab:ablation_retrieval_time}
\end{table}

\begin{table}[t]
    \centering
    \setlength{\tabcolsep}{1.5mm}
    \caption{The Results of Our Methods Using Different Numbers of Re-Ranking Candidates.}
    \vspace{-0.2cm}
    \begin{tabular}{l|cc|cc|cc|cc}
    \toprule
    \multirow{2}{*}{Candidates} &\multicolumn{2}{c|}{Pitts30k-test} & \multicolumn{2}{c|}{MSLS-val} & \multicolumn{2}{c|}{Tokyo24/7} & \multicolumn{2}{c}{Nordland} \\
    \cline{2-9}
    & R@1 & R@5 & R@1 & R@5 & R@1 & R@5 & R@1 & R@5  \\
    \hline
    Top-20 & 94.2 & 97.1 & 93.6 & 97.2 & 97.5 & 98.4 & 96.5 & 98.1 \\
    Top-50 & \textbf{94.4} & 97.3 & 94.3 & 97.7 & \textbf{98.4} & \textbf{99.0} & 97.1 & 98.8 \\
    Top-100 & \textbf{94.4} & 97.5 & \textbf{94.5} & \textbf{98.0} & 98.1 & 98.7 & \textbf{97.2} & 99.0 \\
    Top-200 & \textbf{94.4} & \textbf{97.6} & 94.3 & \textbf{98.0} & 98.1 & \textbf{99.0} & \textbf{97.2} & \textbf{99.1} \\
    \bottomrule
    \end{tabular} 
    \vspace{-0.2cm}
    \label{tab:ablation_candidatesnumber}
\end{table}

\textbf{3) Effect of our novel two-stage paradigm.}
Our novel two-stage VPR paradigm discards the local features used in SelaVPR, and instead only uses global features for both initial retrieval and re-ranking. To demonstrate our re-ranking paradigm, we compare the performance of direct retrieval using binary features and floating-point features, as well as our re-ranking method (binary features for initial retrieval and floating-point features for re-ranking). We conduct experiments using both the DINOv2-base and DINOv2-large backbones with our unified dataset for training. The results are shown in Table \ref{tab:ablation_rerankparadigm}. Since the binary features we use are also low-dimensional, their performance (R@1/R@5/R@10) is significantly lower than that of high-dimensional floating-point features. For the model using DINOv2-base, the R@1 margins on Pitts30k, MSLS-val, Tokyo247, and Nordland are 4.3\%, 4.2\%, 6.3\%, and 16.3\%, respectively. However, in our re-ranking pipeline, the initial search (using binary features) is only expected to return the top-100 candidate images with correct results. The difference between the R@100 results obtained by binary and floating-point features is very slight. After re-ranking the top-100 candidates using high-dimensional floating-point features, the R@1/R@5/R@10 results of our method are not significantly different from those yielded by directly using high-dimensional floating-point features (some are slightly higher, and some are slightly lower).

Since our method uses compact binary features to provide candidate images by calculating Hamming distances, this can further greatly improve retrieval efficiency. Table \ref{tab:ablation_retrieval_time} shows the retrieval latency (including initial retrieval latency and re-ranking latency) of a single query on Pitts30k for three two-stage methods, i.e., TransVPR \cite{transvpr}, SelaVPR \cite{selavpr}, and SelaVPR++. We also provide the latency of direct retrieval with the L2 distance for 4096-dim, 2048-dim, and 512-dim floating-point features, as well as with the Hamming distance for 512-dim binary features. Although our previous work SelaVPR matches local features for re-ranking, it eliminates the time-consuming geometric verification process in the common two-stage method (e.g., TransVPR). Its re-ranking latency is thus more than one order of magnitude lower than that of TransVPR. Our SelaVPR++ directly uses global features for re-ranking, which is further more than two orders of magnitude faster than SelaVPR. As for the initial retrieval latency, once the measurement is determined (e.g., L2 distance), the latency is roughly proportional to the feature dimension. When the feature dimension is 4096, the retrieval latency from 10000 images (i.e., the database of Pitts30k) using L2 distance and the re-ranking latency of SelaVPR (100 candidates) are already within the same order of magnitude. When we reduce the dimension to 512, the retrieval latency also decreases by about 8 times. However, it is still 3.89ms, which is not negligible. In our new re-ranking paradigm, we use the Hamming distance to measure the similarity between binary features, which is more than an order of magnitude faster than calculating the L2 distance between floating-point features of the same dimension (only 0.26ms for 512-dim binary features). Overall, in terms of retrieval latency, while SelaVPR achieves more than an order of magnitude speedup over the common two-stage method, SelaVPR++ achieves more than two orders of magnitude speedup over SelaVPR and 6000× speedup over TransVPR. Compared to the common one-stage methods using 4096-dim global features, SelaVPR++ is also more than 60 times faster.

In addition, although we directly re-rank the top-100 candidates as in other works, it is also feasible to reduce the number of candidates (e.g., 50), which can further reduce latency. Table \ref{tab:ablation_candidatesnumber} shows the results of re-ranking different numbers of candidates. Our method achieves good results in all settings. There is no performance improvement when re-ranking top-200 candidates (compared to top-100), and a certain decrease when re-ranking top-20 candidates (compared to top-50). We believe that re-ranking 50-100 candidates is appropriate.

\begin{table*}[t]
    \centering
    \setlength{\tabcolsep}{1.5mm}
    \caption{The Results of Different Deep Hashing Methods. All Models Are Based on DINOv2-Base and Trained on the GSV-Cities Dataset.}
    \vspace{-0.2cm}
    \begin{tabular}{l|c|c||cc||cc||cc||cc||cc}
    \toprule
    \multirow{2}{*}{Hashing Method} & \multirow{2}{*}{Quantization Loss} & \multirow{2}{*}{STE} &\multicolumn{2}{c||}{Pitts30k-test} & \multicolumn{2}{c||}{MSLS-val} & \multicolumn{2}{c||}{Tokyo24/7} & \multicolumn{2}{c||}{Nordland} & \multicolumn{2}{c}{Average} \\
    \cline{4-13}
    & & & R@1 & R@5 & R@1 & R@5 & R@1 & R@5 & R@1 & R@5 & R@1 & R@5   \\
    \hline
    Direct hashing & $\times$ & $\times$ & 87.9 & 94.9 & 86.4 & 94.3 & 85.7 & 95.6 & 53.2 & 72.4 & 78.3 & 89.3 \\  
    LC hashing \cite{lcdsh} & Pair-wise similarity \cite{lcdsh} & $\times$ & 88.4 & 94.8 & 87.8 & 94.5 & 87.0 & 94.9 & 55.2 & 74.7 & 79.6 & 89.7 \\
    SC hashing & Pair-wise similarity (ours) & $\times$ & \textbf{89.1} & 94.7 & 88.0 & 93.8 & 89.2 & 95.9 & 57.1 & 75.6 & 80.9 & 90.0 \\
    STE \cite{ste} & $\times$ & \checkmark & 88.9 & 95.0 & 88.5 & 94.2 & 88.9 & 96.5 & 58.6 & \textbf{76.8} & 81.2 & 90.6 \\
    SC hashing + STE & Pair-wise similarity (ours) & \checkmark & 89.0 & \textbf{95.3} & \textbf{89.9} & \textbf{94.9} & \textbf{90.2} & \textbf{97.5} & \textbf{58.8} & 76.7 & \textbf{82.0} & \textbf{91.1} \\
    \bottomrule
    \end{tabular} 
    \vspace{-0.2cm}
    \label{tab:ablation_hashing}
\end{table*}

\begin{table*}[!htbp]
    \centering
    \setlength{\tabcolsep}{1.8mm}
    \caption{The Results of Our Method Using the GSV-Cities and Unified Dataset for Training.}
    \vspace{-0.2cm}
    \label{tab:ablation_training_datasets}
    \begin{tabular}{l|c||ccc||ccc||ccc||ccc}
    \toprule
    \multirow{2}{*}{Training Dataset} & \multirow{2}{*}{Backbone} & \multicolumn{3}{c||}{Pitts30k-test} & \multicolumn{3}{c||}{MSLS-val} & \multicolumn{3}{c||}{Tokyo24/7} & \multicolumn{3}{c}{Nordland} \\
    \cline{3-14}
    & & R@1 & R@5 & R@10 & R@1 & R@5 & R@10 & R@1 & R@5 & R@10  & R@1 & R@5 & R@10 \\
    \hline
    GSV-Cities & \multirow{2}{*}{DINOv2-B} & 92.6 & 96.5 & \textbf{97.4} & 92.6 & 96.2 & 97.0 & 96.2 & 97.8 & 98.4 & 81.3 & 91.3 & 94.0 \\
    Unified dataset & & \textbf{93.3} & \textbf{96.6} & \textbf{97.4} & \textbf{94.3} & \textbf{96.9} & \textbf{97.6} & \textbf{96.8} & \textbf{98.4} & \textbf{98.7} & \textbf{94.7} & \textbf{98.3} & \textbf{99.0} \\
    \hline
    GSV-Cities & \multirow{2}{*}{DINOv2-L} & 93.5 & 97.0 & 97.8 & 93.2 & 96.6 & 96.8 & 97.5 & \textbf{99.0} & \textbf{99.7} & 85.2 & 93.4 & 95.7 \\
    Unified dataset & & \textbf{94.4} & \textbf{97.5} & \textbf{98.1} & \textbf{94.3} & \textbf{98.0} & \textbf{98.2} & \textbf{98.1} & \textbf{99.0} & \textbf{99.7} & \textbf{97.0} & \textbf{99.1} & \textbf{99.5} \\
    \bottomrule
    \end{tabular}  
    \vspace{-0.2cm}
\end{table*}

\textbf{4) Effect of our similarity-constrained deep hashing.} Our deep hashing method combines the similarity-constrained loss and straight-through estimation (STE) \cite{ste}. The former is inspired by the locality-constrained deep hashing \cite{lcdsh}, which preserves pair-wise similarity (inner product) between images and uses a sigmoid transformation on the inner product. However, our work directly constrains the cosine similarity of feature pairs before and after quantization without using the sigmoid function. In this subsection, we compare the performance of direct hashing, locality-constrained (LC) hashing, STE, our similarity-constrained (SC) hashing, and our complete method (i.e., SC hashing + STE). To widen the performance gap between different methods, all models are based on the DINOv2-base backbone (outputting 512-dim features) and trained only on GSV-Cities. The results are shown in Table \ref{tab:ablation_hashing}. We first summarize the results in terms of the average R@1/R@5 performance. The direct hashing method has the worst result because it uses the floating-point features when training the model and directly quantizes these features to obtain binary features during inference, which causes obvious information loss (or a gap between training and inference). LC hashing improves performance on all datasets by constraining the similarity of feature pairs before and after quantization, while our SC hashing is not only more concise than LC hashing but also achieves further improvement on the VPR task. Even so, STE achieves better overall performance than these methods by directly solving the gradient issue and using quantized binary features to calculate the metric loss for end-to-end training. Our complete method combines SC hashing and STE, and finally achieves the best performance among these methods. In addition, as for R@1 results on individual datasets, apart from our complete method, SC hashing achieves the best results on urban datasets (Pitts30k and Tokyo24/7) but is inferior to STE on suburban and natural datasets (MSLS and Nordland). Our approach yields outstanding results in different scenes simultaneously by combining these two methods.

\textbf{5) Effect of unifying and merging training datasets.}
The main contribution of the training strategy in our SelaVPR++ is to unify and merge several common VPR training datasets with the same training protocol for model training. To validate its effectiveness, we compare the performance of models trained with only GSV-Cities \cite{gsv} and our proposed unified dataset. We conduct two sets of experiments that use both DINOv2-base and DINOv2-large backbones and produce 2048-dim and 4096-dim features respectively (as above). The results are shown in Table \ref{tab:ablation_training_datasets}, and we summarize two points from it. \textbf{(1)} The model trained on the unified dataset consistently outperforms the one only using the GSV-Cities dataset on all evaluation datasets, regardless of whether DINOv2-base or DINOv2-large is used as the backbone. This demonstrates that training with our unified dataset can enhance the robustness and achieve better recognition performance. \textbf{(2)} The magnitude of performance improvement varies on different test datasets, with the largest gain achieved on Nordland (composed of natural landscapes), followed by MSLS-val (containing both urban and suburban scenes), and the smallest gains on Pitts30k and Tokyo24/7 (primarily consisting of urban street view images). For example, when using DINOv2-base, the R@1 improvements on Pitts30k, Tokyo24/7, MSLS-val, and Nordland are 0.7\%, 0.6\%, 1.7\%, and 13.4\%, respectively. This is because the GSV-Cities training set mainly consists of urban images, while our unified dataset covers more diverse VPR scenes. As a result, our model trained on the unified dataset is highly robust against diverse challenges, thus performing better and more consistently in various scenarios. 

\begin{figure*}[t]
    \centering
    \includegraphics[width=0.85\linewidth]{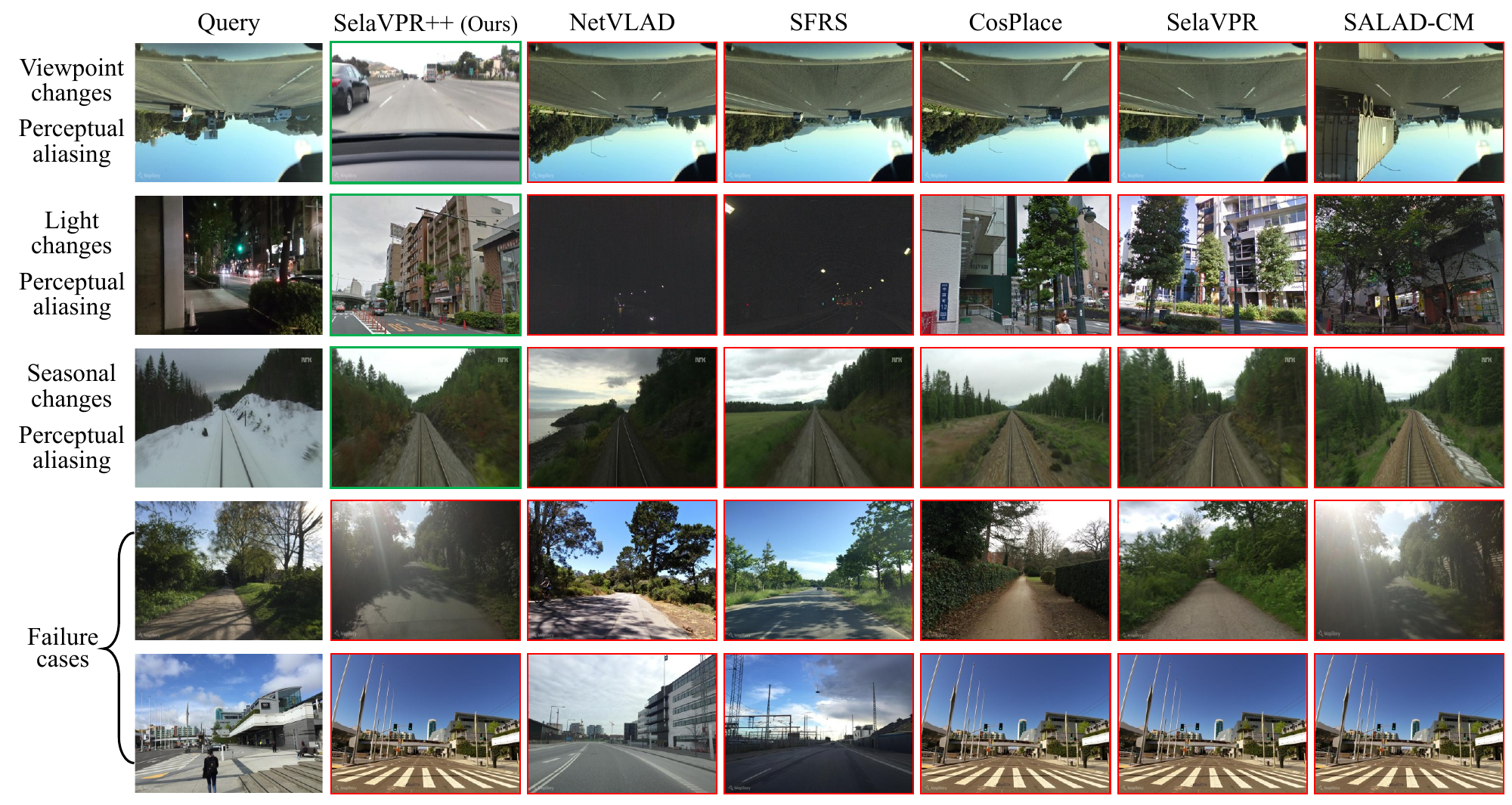}
    \vspace{-0.3cm}
    \caption{
    Some examples of corner cases (success cases in the first three and failure cases in the last two). For the first example, the query is rotated 180$^\circ$. All other methods return the inverted but wrong results, while only our SelaVPR++ gets the correct image (with normal orientation). The second example shows a nighttime query with weak light and limited visibility. Our method is highly robust against light changes and gets the right database image (exhibiting obvious viewpoint changes with the query), but others fail. The third and fourth query images are taken in nature, lacking discriminative landmarks and buildings. Our method, which can capture other valuable details (e.g., terrain characteristics and vegetation), succeeds in the third case. However, this does not guarantee a correct match every time, especially when the plants shown in the query image lack distinctive characteristics and occupy a majority of the image region (i.e., the fourth case). In the last example, most methods return the same result that is geographically close to the query but exceeds the predefined threshold (i.e., failed). Although SALAD-CM introduces a novel CliqueMining strategy explicitly tailored to address this issue, it still yields the same wrong result as ours.
    }
    \vspace{-0.4cm}
    \label{resultfig_suppl}
\end{figure*}

\section{Discussion}
Discussion on Efficiency: In addition to excellent recognition performance, high efficiency is another major advantage of our method compared to other methods. The efficiency advantages of SelaVPR++ can be highlighted as follows: \textbf{(1)} Our model adaptation method is a parameter-, time-, and memory-efficient fine-tuning method. Compared with full fine-tuning, it has only 8.7\% of the trainable parameters, consumes 11.5\% of the GPU memory, and less than half of the training time for the DINOv2-large backbone. Meanwhile, it is also more efficient than the partial fine-tuning method (less than half of trainable parameters and GPU memory usage, as well as less training time for DINOv2-base). \textbf{(2)} Our novel two-stage VPR method is efficient in retrieval time. It achieves more than 6000× speedup over TransVPR and 60× speedup over the one-stage method with 4096-dim global descriptors. With both excellent performance and high efficiency, SelaVPR++ has addressed most issues for real-world large-scale VPR.

Discussion on corner cases: Although SelaVPR++ is highly robust against most challenges in VPR, there are a few corner cases that it struggles to tackle effectively. Fig. \ref{resultfig_suppl} shows some examples of corner cases. For some extreme viewpoint changes or illumination changes (or even both simultaneously), other methods fail while ours shows good robustness and yields correct results. Besides, when the query image taken at natural scenes lacks discriminative landmarks, our method can still capture other useful details to identify places. However, our method fails in a few natural scenes. One promising solution is to increase the descriptor dimension to provide more detailed information, which requires making a reasonable trade-off according to our needs. In another failure case, the retrieved database image and the query are collected from two close locations and represent almost the same content, but the geographical distance between them exceeds the set threshold. This is considered a wrong match in VPR and is encountered by all methods. We believe that a direct mitigation way is to increase the geographical density of image collection when building the database.

\section{Conclusions}
In this paper, we extend our previous conference work SelaVPR, and propose SelaVPR++, which achieves significant improvements in recognition performance and efficiency. Initially, we design a parameter-, time- and memory-efficient MultiConv adaptation method to seamlessly adapt pre-trained foundation models for the VPR task. The feature representation produced by the adapted foundation model is more focused on discriminative landmarks to differentiate places, thus bridging the gap between the tasks of model pre-training and VPR. Then, we propose an innovative two-stage VPR paradigm that uses compact binary features for initial retrieval and robust floating-point (global) features for re-ranking, thereby greatly improving retrieval efficiency. Furthermore, we improve our training strategy and merge several commonly used VPR datasets in a unified training protocol for training more robust VPR models. The experimental results demonstrate that our SelaVPR++ is efficient in trainable parameters, memory usage, as well as (training and retrieval) time. It also outperforms previous SOTA methods on VPR benchmark datasets by a considerable margin. We believe that our SelaVPR++ has addressed most challenges in the VPR task and paved the way for real-world large-scale VPR applications.

\section*{Acknowledgments}
This work was supported by the National Key R\&D Program of China (2022YFB4701400/4701402), SSTIC Grant (KJZD20230923115106012, KJZD20230923114916032, GJHZ20240218113604008), the Key Project of Pengcheng Laboratory, and National Natural Science Foundation of China (62402252).

\bibliographystyle{IEEEtran}
\bibliography{IEEEabrv, reference}

\vfill

\end{document}